\documentclass[journal]{IEEEtran}
\usepackage{graphicx}                                % some people may need to use [dvips] option
\usepackage{amsmath}
\usepackage{multirow}
\usepackage{algorithm}
\usepackage{subfigure}
\usepackage{algorithmic}
\usepackage{cite}
\usepackage{amsthm}
\usepackage{amssymb}
\usepackage{array}
\usepackage{setspace}
\usepackage{booktabs}
\usepackage{float}
\usepackage{threeparttable}

\begin{document}
\title{Unsupervised Self-training Algorithm Based on Deep Learning for Optical Aerial Images Change Detection}

\author{Yuan Zhou,~\IEEEmembership{Senior Member,~IEEE,}
        Xiangrui Li % <-this % stops a space
% <-this % stops a space
\thanks{Yuan Zhou is with the School of Electrical and Information Engineering, Tianjin University, Tianjin 300072, China. (e-mail: zhouyuan@tju.edu.cn)}% <-this % stops a space
\thanks{Xiangrui Li is with Tianjin International Engineering Institute, Tianjin University, Tianjin 300072, China.}
}


\maketitle

\begin{abstract}
Optical aerial images change detection is an important task in earth observation and has been extensively investigated in the past few decades. Generally, the supervised change detection methods with superior performance require a large amount of labeled training data which is obtained by manual annotation with high cost. In this paper, we present a novel unsupervised self-training algorithm (USTA) for optical aerial images change detection. The traditional method such as change vector analysis is used to generate the pseudo labels. We use these pseudo labels to train a well designed convolutional neural network. The network is used as a teacher to classify the original multitemporal images to generate another set of pseudo labels. Then two set of pseudo labels are used to jointly train a student network with the same structure as the teacher. The final change detection result can be obtained by the trained student network. Besides, we design an image filter to control the usage of change information in the pseudo labels in the training process of the network. The whole process of the algorithm is an unsupervised process without manually marked labels. Experimental results on the real datasets demonstrate competitive performance of our proposed method.
\end{abstract}

\begin{IEEEkeywords}
Change detection, convolutional neural network, unsupervised learning, optical aerial images.
\end{IEEEkeywords}

\IEEEpeerreviewmaketitle

\section{Introduction}

\IEEEPARstart{C}{hange} detection in remote sensing is the process of detecting the changes that occurred on the surface of the Earth by analyzing two images obtained at different times over the same geographical area\cite{radke2005image,dominguez2018a}. It has provided great convenience for disaster assessment\cite{1294467}, agricultural evaluation\cite{radke2005image}, urban expansion\cite{manonmani2010remote} and so on. With the rapid development of aviation technology, especially unmanned aerial vehicles, a large number of aerial images which is a kind of remote sensing images with high ground resolution and rich content can be acquired\cite{liu2019coastal}. Aerial images have become one of the main data sources for change detection in the field of remote sensing. In this paper, we focus on change detection for optical aerial images.

In the past few decades, a variety of change detection techniques have emerged driven by great social and economic benefits. In the literature, the mainstream methods generally compare the multitemporal images to obtain a difference image (DI) and the analysis will be based on the DI to get the change detection result\cite{article}. The threshold and clustering methods which have good performance in classification tasks are usually adopted for the analysis of the DI\cite{kittler1986minimum,yetgin2012unsupervised,krinidis2010a,gong2012change}. At present, the core of change detection methods are how to obtain a quality DI because the DI contains the key change information of the multitemporal images and it is the basis of subsequent operations.

To generate a DI, image differencing\cite{quarmby1989monitoring}, image ratioing\cite{howarth1981procedures} are the intuitive and straightforward techniques which perform algebraic operations on the pixels of the multitemporal images. Change vector analysis (CVA)\cite{bovolo2007a} compares each pair of corresponding pixels of the two multichannel images to get the change vector whose magnitude and direction represent the intensity and direction of the change, respectively. Principal component analysis (PCA)\cite{deng2008pca-based} transforms the images with high-dimension into a low-dimensional feature space which contains the key information to reduce redundancy during the generation of the DI. Multivariate alteration detection (MAD)\cite{nielsen1998multivariate} is a multivariate statistical analysis method that uses the correlation between comprehensive component pairs to reflect the overall correlation between two sets of components, eliminating the adverse effects of correlation between channels on change detection. Iteratively reweighed multivariate alteration detection (IR-MAD)\cite{nielsen2007the} is the extension of MAD. Different observed images have different weights in IR-MAD. Large weights will be assigned to observations which show little change and observations which show large change will have little weights.

However, the typical methods mentioned above are weak in feature representations, so it can not obtain enough ideal result in the change detection task of optical aerial images with high ground resolution and rich image content. In recent years, deep neural networks (DNNs) have received extensive attention as a tool for extracting features of data. A variety of DNNs have acquired good performance in many tasks including change detection, such as deep belief networks (DBNs)\cite{hinton2006a}, sparse autoencoders (AEs)\cite{6910306}, recurrent neural network (RNNs)\cite{hochreiter1997long} and convolutional neural networks (CNNs)\cite{article2}. According to whether the hand-marked labels is needed, the methods for change detection based on DNNs can be divided into supervised and unsupervised methods.

Supervised methods need the hand-marked labels to train the models. In\cite{zhan2017change}, a deep siamese convolutional neural network with the weighted contrastive loss\cite{1640964} was proposed for aerial remote sensing image change detection. Zhang et al.\cite{zhang2019triplet-based} proposed to use a improved triplet loss function\cite{7298682} to train a deep siamese semantic network which have a good performance. Three types of fully convolutional neural network (FCNN) were proposed to perform change detection on multitemporal images in\cite{8451652}. Mou et al.\cite{mou2019learning} proposed a novel recurrent convolutional neural network (ReCNN) which combines recurrent neural network and convolutional neural network to learn a spectral-spatial-temporal feature representation for change detection in multispectral images. These supervised methods which rely on the labels for training have a good performance. However, the labels are obtained by manual annotation with high cost. Therefore, the supervised methods is not practical in many conditions.

Unsupervised methods which can generate the change detection result without the hand-marked labels have been widely used in practice. In\cite{liu2018a}, a symmetric convolutional coupling network (SCCN) transformed the input images into a consistent feature space, where the DI can be obtained. The parameters of the network can be optimized by a coupling function without the hand-marked labels. In\cite{gao2016automatic}, a PCANet which uses PCA filters as convolutional filters was adopted as the classification model for change detection. The log-ratio algorithm is used to generate the pseudo labels. The pixels which have high probability of being changed or unchanged will be select by Gabor wavelets and FCM algorithm to train the PCANet. Gong et al.\cite{gong2017generative} proposed to use generative adversarial networks (GAN) to carry on the unsupervised change detection. The typical method, such as CVA, is used to generate the pseudo labels, then the small training patches are sampled from the pseudo labels by a predetermined standard\cite{gong2016change}. The final result can be obtained by the trained GAN’s generator. Liu et al.\cite{liu2020convolutional} applied a convolutional neural network based on transfer learning (TLCNN) to change detection for optical aerial images. The network is pretrained on an open labeled dataset which is used for the supervised semantic segmentation task. The log-ratio algorithm is applied to generate the pseudo labels. Then the network is further trained by utilizing the change informations extracted by the pretrained model and contained in the pseudo labels.

To sum up, most unsupervised methods take the pseudo labels generated by the typical methods as the foundation of subsequent operations. The effect of the algorithm depends on the quality of the pseudo labels. However, the pseudo labels generated by the typical methods often don’t have a good quality. The errors in the pseudo labels may influence the final result. Therefore, it is necessary to reduce the influence of the wrong change information in the pseudo labels. Most existing methods for reducing the impact of errors in the pseudo labels are through sampling small patches whose central pixel has high probability of being changed or unchanged to train the model \cite{gao2016automatic,gong2017generative,gong2019a}, or utilizing knowledge from the additional labeled dataset\cite{liu2020convolutional,yang2019transferred}. The first type of methods ignore the spatial feature of the multitemporal images. For the second type of methods, difference between datasets may introduce additional errors and the final training is still guided by the original pseudo labels.

To address these issues, in this paper, we propose a novel unsupervised self-training algorithm (USTA) for optical aerial images change detection. A CNN based on U-Net\cite{ronneberger2015u-net:} is designed as the subject of training in our method. The unsupervised self-training strategy is proposed, which has three main steps: 1) the designed network is trained with the pseudo labels generated by the typical methods, 2) the trained network is used as a teacher to classify the original optical aerial images to generate another set of pseudo labels, 3) a student network with the same structure as the teacher is jointly trained with the two set of pseudo labels. Then the final change detection result can be obtained by the trained student network. In addition, we design an image filter to control the usage of change information in the pseudo labels during the training process of the models. In our method, the training samples can be large images instead of small patches, so the spatial features of the multitemporal images can be well utilized. Besides, the whole process can be completed on a single optical aerial images dataset. The experiment results show that our method can suppression the wrong information in the pseudo labels very well and has a better performance for optical aerial images change detection compared with the state-of-the-art methods.

The novel contributions of this paper can be concluded in three aspects as follows.
\begin{enumerate}
\item[1)]We design a CNN based on U-Net. The network can extract high-level features of two input images efficiently and accurately, then the DI can be output by deal with the features automatically.
\item[2)]We design an image filter to control the usage of change information in the pseudo labels. The correct and wrong information can be filter out, which will be given more and less weight in the training, respectively.
\item[3)]We propose a novel training strategy named unsupervised self-training. The usage of the jointly pseudo labels can reduce the negative influence of errors in the single set of pseudo labels.
\end{enumerate}

The rest of this letter is organized as follows. Section II shows the relevant work of the proposed method. In Section III, we describe the details of the proposed algorithm. In Section IV, we analyze the performance of the proposed method in the experiments. Section V discusses the influence of the related factors on the change detection results. In the end, Section VI concludes the paper.

\begin{figure*}
\centering
\includegraphics[width=0.9\linewidth]{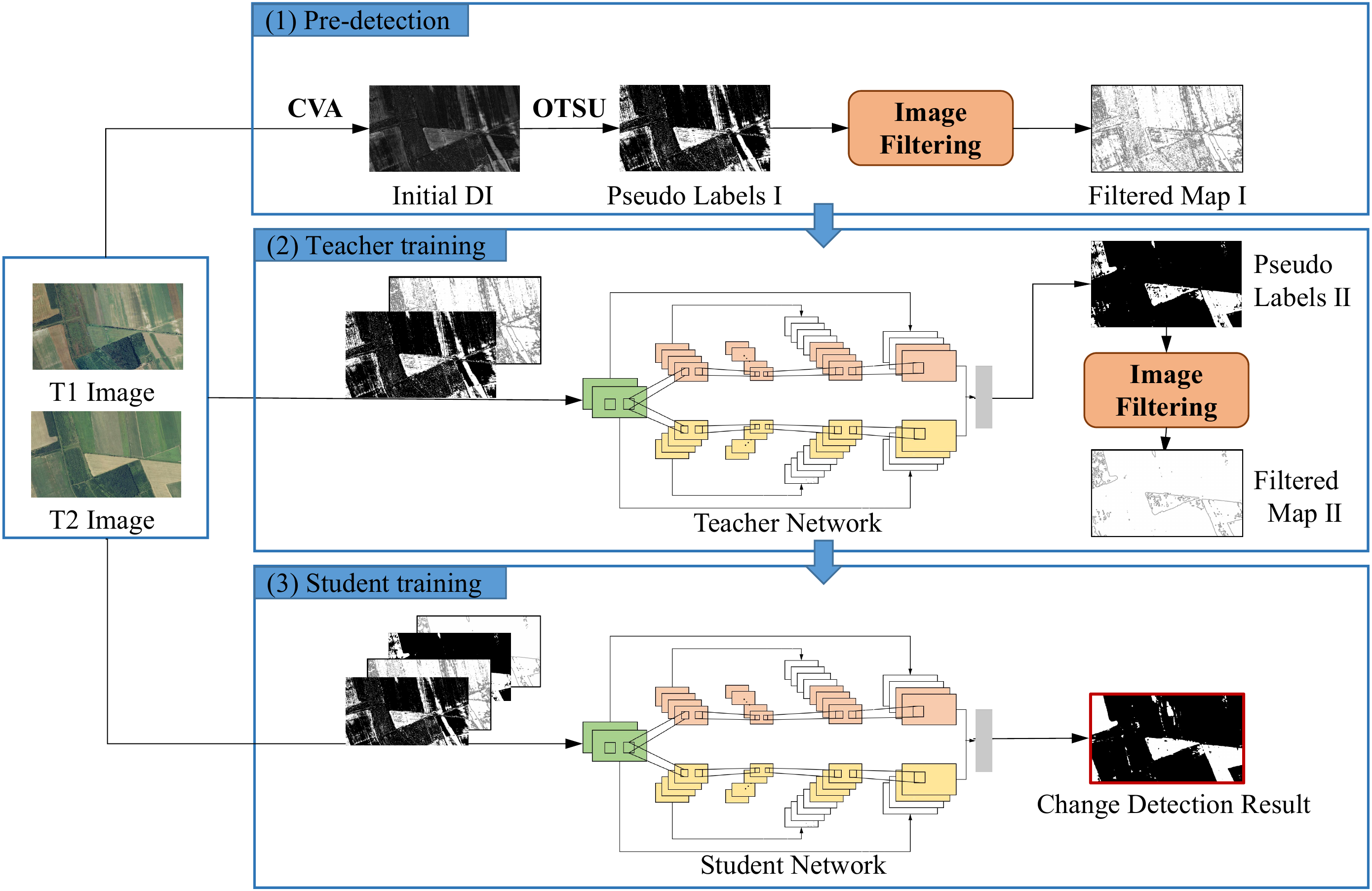}
%\caption{fig1}
\caption{The overall process of the proposed unsupervised self-training algorithm (USTA). 1) In the pre-detection module, traditional methods are used to detect the multitemporal images to obtain initial change detection result as the pseudo label I. The designed image filter is used to obtain the filtered map I, 2) in the teacher training module, the designed CNN as a teacher is trained on the pseudo label I and filtered map I. The trained teacher network can generate another change detection result as the pseudo label II, then the filter is used to generate the corresponding filtered map II, 3) in the student training module, the joint pseudo labels and filtered maps obtained in the first two modules are used to train another student network. The final change detection result is obtained by the trained student network.}
\label{overall}
\end{figure*}

\section{Related Work}

\subsection{U-Net}
U-Net is a convolutional neural network proposed for biomedical image segmentation in\cite{ronneberger2015u-net:}. The network architecture consists of a contracting path and an expansive path. The size of the feature map is halved by pooling layers and the number of feature channels is doubled by convolution layers in every step of the contracting path where the high-level feature can be extracted. Every step in the expansive path utilize an upsampling to double the size of the feature map. The result of the upsampling is concatenated with the corresponding feature map from the contracting path, then the number of fusional feature channels is halved by convolution layers. The skip connection in U-Net improve the network's effect of mining features. The segmentation result can be obtained in the end of the expansive path.

\subsection{Image Filtering}
Image filtering is an important technique in image processing such as image denoising, image enhancement and edge detection. It is a neighborhood operation that uses the values of pixels around a given pixel to determine the final output value of this pixel. Through filtering, some features can be emphasized and some unwanted parts of the image can be removed. A variety of filters have been widely used such as harmonic mean filter, median filter, max filter and Gaussian filter. In this paper, we design a novel filter which is used to filter out the correct and wrong information in the pseudo labels.

\subsection{Self-training}
Self-training is a semi-supervised learning strategy which has shown success for several tasks\cite{1053799,10.3115/981658.981684,riloff2003learning}. Self-training first uses labeled data to train a model. The model is used to label unlabeled data and then the labeled data and selected unlabeled data with high confidence are used to jointly train a new model. The architectures for the models trained by different label can be the same or different. Although Self-training have a good performance in some semi-supervised tasks, it still need several labeled data to train the initial model. Besides, it is not suitable for pixel-level classification tasks such as change detection because the selection of the reliable pixel samples is difficult. In order to meet the need of no label in optical aerial images change detection, in this paper, we propose a unsupervised self-training strategy.

\section{Methodology}

In this section, the concrete implementation of our method is described in detail. The overall process which consists of three modules is shown in Fig. \ref{overall}: 1) in the pre-detection module, CVA and Otsu\cite{otsu1979a} methods are used to deal with the multitemporal images to obtain initial change detection result as the pseudo label I, then we use the designed image filter on the pseudo label I to obtain the filtered map I, 2) in the teacher training module, the well designed CNN as a teacher will be trained on the pseudo label and filtered map obtained in the pre-detection module. The trained teacher network can generate another change detection result as the pseudo label II based on the original multitemporal images, then the filter is used to generate the corresponding filtered map II, 3) in the student training module, we use the joint pseudo labels and filtered maps obtained in the first two modules to train another student CNN with the same architecture as the teacher. After training, the student network can generate the final change detection result with high quality. The detailed descriptions of the designed filter, the CNN and the unsupervised self-training strategy are reported are reported in the subsection A-C.

\subsection{Designed Image Filter}
A serious challenge in our unsupervised self-training algorithm is how to deal with the noise in the pseudo labels. The change map as the pseudo labels generated by the typical methods or the teacher network generally contains some pixels which are classified wrongly. Each erroneous pixel is a noisy sample. There is a high chance that noisy samples mislead and confuse the network during training. The impact of the erroneous pixels is particularly serious during the training of the teacher network. The pseudo labels generated by the typical methods generally don’t have a good quality. The teacher network will be terrible if these pseudo labels are used directly for training. Besides, the quality of the teacher network determines the quality of the pseudo label II which will be used in the student network training. Considering the negative influence of noise in the pseudo label, we design an image filter to address this issue.

\begin{figure}
\centering
\includegraphics[width=\linewidth]{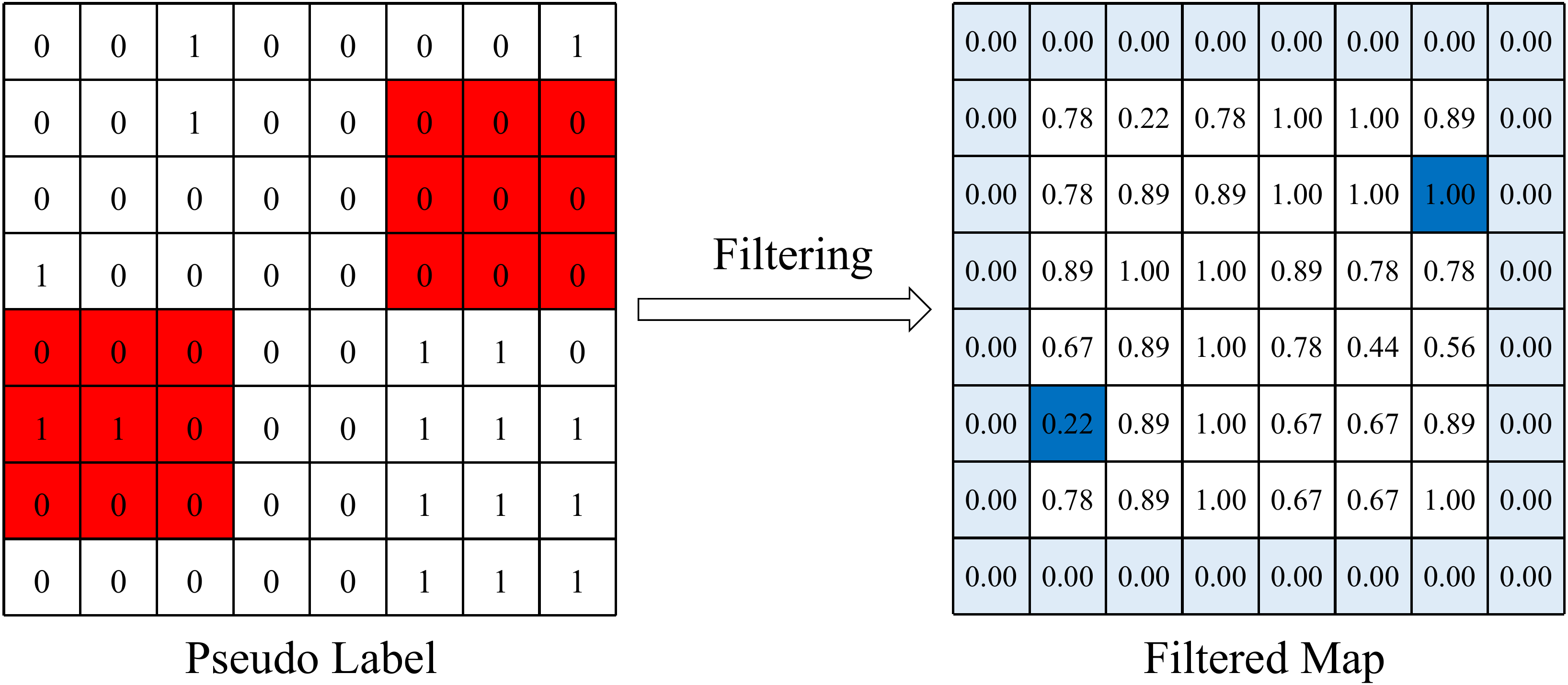}
\caption{A intuitive exhibition of the filtering process. The filtering result of the red region in the pseudo label is corresponding to the dark blue in the filtered map. 0 is padded to the the edges of the filtered map to ensure the pixels of the image before and after filtering are in one-to-one correspondence.}
\label{filter}
\end{figure}

Suppose that $\textbf{CM}=\{\textup{cm}(\textit{i},\,\textit{j}), \textup{1}\leq \textit{i}\leq \textit{H}, \textup{1}\leq \textit{j}\leq \textit{W}\}$ represents a pseudo label generated by the typical methods or the teacher network with a size of $\textit{H} \times \textit{W}$. $\textup{cm}(\textit{i},\,\textit{j})$ represents the pixel in row $\textit{i}$ and column $\textit{j}$ of the pseudo label. Let $\textbf{T}=\{\textup{0},\textup{1}\}$ be the set of labels. 0 and 1 denote the unchanged class and changed class, respectively. For a label pixel $\textup{cm}(\textit{i}, \,\textit{j})$, we filter the neighborhood with a center at the position $(\textit{i}, \,\textit{j})$ and of odd-size $\textit{w}\times \textit{w}$, which can be described as follows:
\begin{equation}
\textup{pc}(\textit{i}, \,\textit{j})=\sum_{\textit{l}=-\frac{\textup{1}}{\textup{2}}(\textit{w}-\textup{1})}^{\frac{\textup{1}}{\textup{2}}(\textit{w}-\textup{1})} \textup{cm}(\textit{i}+\textit{l}, \textit{j}+\textit{l}) \odot \frac{\textup{cm}(\textit{i}, \,\textit{j})}{\textit{w} \times \textit{w}}
\end{equation}
where $\textup{pc}(\textit{i}, \,\textit{j})$ is the result of filtering for a neighborhood. $\odot$ represents exclusive NOR (XNOR), which is a mathematical operator applied to logical operations. If two pixels have the same label, the result of $\odot$ on them is 1 and if two pixels have the different labels, the result of $\odot$ on them is 0. $\textit{l}$ is the offset of the position coordinates. $\textup{cm}(\textit{i}+\textit{l}, \textit{j}+\textit{l})$ represent pixels at different positions on the neighborhood. Each pixel in the neighborhood will been done exclusive NOR operation with the central pixel $\textup{cm}(\textit{i},\,\textit{j})$. $\textup{pc}(\textit{i},\,\textit{j})$ is the proportion of the pixels with the same value as the central pixel to the total pixels in the neighborhood, which can be seen as the probability that the central pixel $\textup{cm}(\textit{i},\,\textit{j})$ is the correct label. If a pixel has a neighborhood in which most pixels have the same labels with it, we consider the probability that this pixel is the correct label is high. As shown in Fig. \ref{filter}, the filter is used to deal with the neighborhood of size $\text{3}\times \text{3}$. $\textup{cm}(\text{3},\text{7})$ in the red area at the upper right corner of the pseudo label has a neighborhood in which all pixels have the same labels with it, so it has very high probability of being a correct label. $\textup{cm}(\text{6},\text{2})$ has a neighborhood in which only one pixel have the same label with it, so it has very low probability of being a correct label.

For pixels at the edges in the pseudo label, we can't find a neighborhood surrounding it. In order to ensure that the pixels of the image before and after filtering are in one-to-one correspondence, we pad 0 to the edges of the filtered map. The pseudo labels and filtering results we obtained in practical generally have large sizes, so the padding pixels hardly affect subsequent operations. Finally, the filtering result $\textbf{PC}=\{\textup{pc}(\textit{i}, \,\textit{j}), \textup{1}\leq \textit{i}\leq \textit{H}, \textup{1}\leq \textit{j}\leq \textit{W}\}$ on $\textbf{CM}$ can be obtained.

Through the filtering, the pixels with low and high probability of being correct labels are filtered out. Compared with the methods which crop reliable small patches from the pseudo labels\cite{gao2016automatic,gong2017generative,gong2019a}, the spatial features of the multitemporal images are well reserved. We can utilize the filtering result to control the usage of pixels of the pseudo labels during the process of training the network to reduce the adverse effects of the noise. The details of training will be described in Section III-C.

\subsection{Change Detection Network}
\begin{figure*}
\centering
\includegraphics[width=0.9\linewidth]{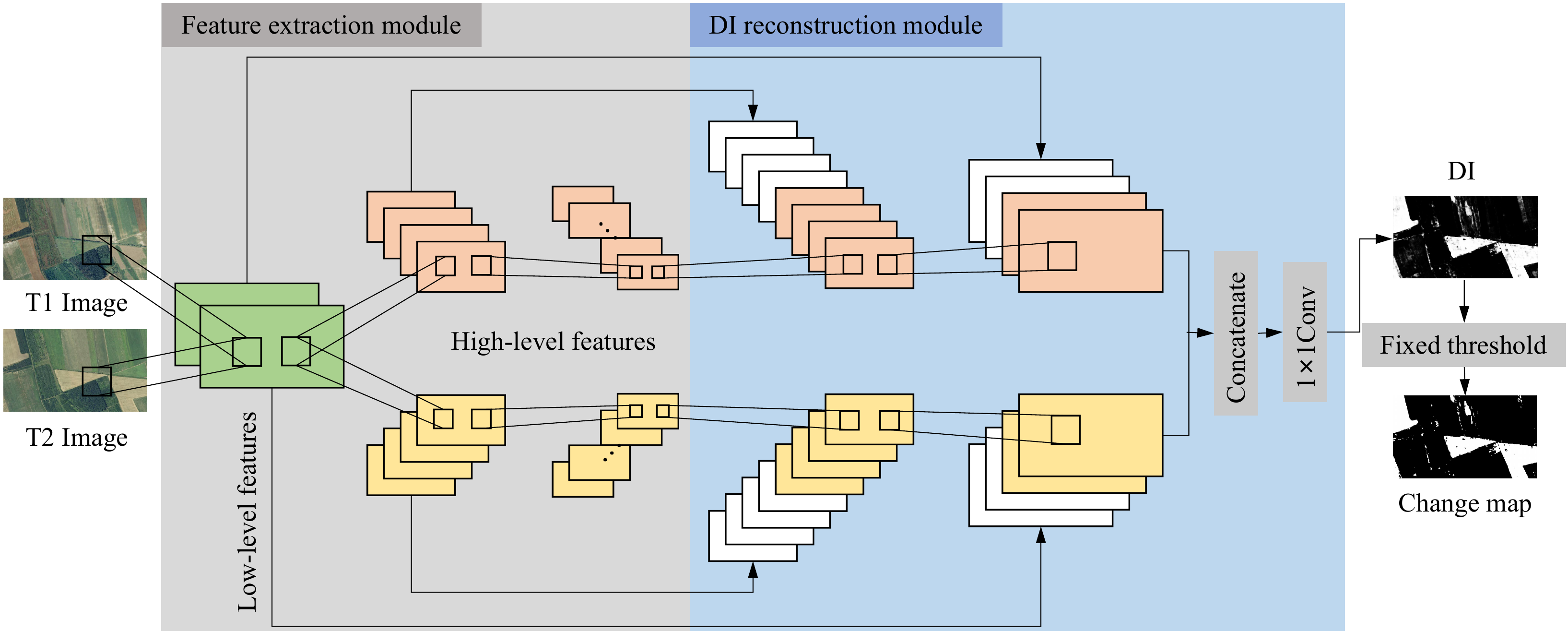}
\caption{The architecture of the change detection network. Low-level and high-level features can be extracted in the feature extraction module. The DI can be automatically reconstruction through the reconstruction module.}
\label{model}
\end{figure*}

In this subsection, We will describe the details of the designed CNN. The network architecture consists two parts as shown in Fig. \ref{model}: a feature extraction module and a DI reconstruction module.

The effect of feature extraction is related to the number of layers of the network. The shallow layers extract low-level features which are highly versatile such as edges and textures. As the number of layers continues to increase, high-level semantic features which are more abstract can be extracted\cite{zeiler2014visualizing}. In our task for optical aerial images change detection, the two images acquired at different times over the same geographical area are homogeneous which means they are captured by the same type of optical sensors and have similar low-level characteristics. Therefore, we use the shallow layers of the feature extraction module to deal with the two images simultaneously and the low-level features are extracted in the same way. The complexity of the network and the requirements of training for the amount of training data can be reduced to avoid overfitting\cite{ioffe2015batch} in this way. Then the high-level semantic features of the two images are separately extracted by the two branches networks with the same structure but different parameters of the feature extraction module. Layers of the module, size and number of convolution kernels are shown in Table \ref{Architecture of the Change Detection Model}. The DConv layer contains two continuous Convolution-BatchNorm-ReLu operations. DConv1-Pooling3 layers are used to extract low-level features. DConv4-Pooling5 layers are the architecture of the two branches networks used to extract high-level features of the two images separately. The number of feature channels is increased by convolution layers and the size of the feature map is halved by pooling layers.

In the DI reconstruction module, the features of the two images extracted by the feature extraction module are separately dealt with by two branches networks with the same structure but different parameters. The configurations of the branch are shown in Table \ref{Architecture of the Change Detection Model}. The TConv layer contains a transposed convolution\cite{pan2016shallow} which double the size of the feature map. We utilize the skip connection in U-Net\cite{ronneberger2015u-net:} to improve the network's ability of mining features. The result of the transposed convolution is concatenated with the feature map of the same size from the feature extraction module. Then the number of fusional feature channels is halved by convolution layers. In the end of the two branches, the feature maps have been reconstructed to the same size as the original images. After concatenating the feature maps output by the two branches, to obtained the DI, we use 1$\times$1 convolution to reduce the number of channels to 1, and Sigmoid function as the final activation function. The change detection result can be obtained by a simple fixed threshold method on the DI.

The high-level features can be extracted efficiently and accurately by the the unique feature extraction method. The DI can be automatically reconstruction through the reconstruction module. It is worth mentioning that we do not use fully connected layers which fixes the size of the input images. Therefore, remote sensing images of different sizes can be detected directly through our proposed network.

\begin{table}[]
\centering
\renewcommand{\arraystretch}{1.1}
\caption{Configuration of the Change Detection Network}
\label{Architecture of the Change Detection Model}
\begin{tabular}{cccccc}
\hline
\multicolumn{3}{c}{Feature extraction module} & \multicolumn{3}{c}{DI reconstruction module} \\
Layer           & K-Size               & K-Num       & Layer          & K-Size               & K-Num       \\ \hline
DConv1          & 3$\times$3           & 64          & TConv6         & 2$\times$2           & 512         \\
Pooling1        & 2$\times$2           & -           & DConv6         & 3$\times$3           & 512         \\
DConv2          & 3$\times$3           & 128         & TConv7         & 2$\times$2           & 256         \\
Pooling2        & 2$\times$2           & -           & DConv7         & 3$\times$3           & 256         \\
DConv3          & 3$\times$3           & 256         & TConv8         & 2$\times$2           & 128         \\
Pooling3        & 2$\times$2           & -           & DConv8         & 3$\times$3           & 128         \\
DConv4          & 3$\times$3           & 512         & TConv9         & 2$\times$2           & 64          \\
Pooling4        & 2$\times$2           & -           & DConv9         & 3$\times$3           & 64          \\
DConv5          & 3$\times$3           & 1024        & Conv10         & 1$\times$1           & 16          \\
-               & -                    & -           & Conv11         & 1$\times$1           & 1           \\
-               & -                    & -           & Sigmoid        & -                    & -           \\ \hline
\end{tabular}
\end{table}

\subsection{Unsupervised Self-training Strategy}
In this subsection, we will elaborate on the entire learning process of our unsupervised self-training strategy.

\subsubsection{Pre-detection}
In the pre-detection stage as shown in Fig. \ref{overall}, two optical aerial remote sensing images $\textbf{X}_{\textup{1}}$ and $\textbf{X}_{\textup{2}}$ are considered. $\textbf{X}_{\textup{1}}=\{\textit{x}_{\textup{1}}(\textit{i},\,\textit{j}), \textup{1}\leq \textit{i}\leq \textit{H}, \textup{1}\leq \textit{j}\leq \textit{W}\}$ and $\textbf{X}_{\textup{2}}=\{\textit{x}_{\textup{2}}(\textit{i},\,\textit{j}), \textup{1}\leq \textit{i}\leq \textit{H}, \textup{1}\leq \textit{j}\leq \textit{W}\}$ with a size of $\textit{H}\times \textit{W}$ are obtained at different times $\textit{t}_{\textup{1}}$ and $\textit{t}_{\textup{2}}$ over the same geographical area, respectively. CVA is used to deal with the two images to generate the $\textbf{DI}_{\textup{1}}=\{\textup{di}_{\textup{1}}(\textit{i},\,\textit{j}), \textup{1}\leq \textit{i}\leq \textit{H}, \textup{1}\leq \textit{j}\leq \textit{W}\}$. As a thresholding algorithm, Otsu has the ability to determine the threshold automatically; it has been widely used in change detection with the advantages of easy implementation and good performance. We use Otsu to classify the pixels of the $\textbf{DI}_{\textup{1}}$ into changed and unchanged classes and we can obtain a change map $\textbf{CM}_{\textup{1}}=\{\textup{cm}_{\textup{1}}(\textit{i},\,\textit{j}), \textup{1}\leq \textit{i}\leq \textit{H}, \textup{1}\leq \textit{j}\leq \textit{W}\}$ as the pseudo label I. We use the designed filter to deal with the pseudo label $\textbf{CM}_{\textup{1}}$. The filtered map I is represented by $\textbf{PC}_{\textup{1}}=\{\textup{pc}_{\textup{1}}(\textit{i},\,\textit{j}), \textup{0}\leq \textit{i}\leq \textit{H}, \textup{1}\leq \textit{j}\leq \textit{W}\}$, where $\textup{0}\leq \textup{pc}_{\textup{1}}(\textit{i},\,\textit{j})\leq \textup{1}$. $\textbf{PC}_{\textup{1}}$ will be used to control the usage of pixels in $\textbf{CM}_{\textup{1}}$ in the training process of the teacher network.

\subsubsection{Teacher Training}
In the teacher network training stage as shown in Fig. \ref{overall}, given two optical aerial remote sensing images $\textbf{X}_{\textup{1}}$ and $\textbf{X}_{\textup{2}}$, $\textbf{DI}_{\textup{2}}=\{\textup{di}_{\textup{2}}(\textit{i},\,\textit{j}), \textup{1}\leq \textit{i}\leq \textit{H}, \textup{1}\leq \textit{j}\leq \textit{W}\}$ is the output of the teacher model, which can be defined as follows:
\begin{equation}
\textbf{DI}_{\textup{2}}=\textit{f}_{\theta_{\textit{r}}}^{\,\textup{teacher}}\left(\textit{g}_{\theta_{\textit{e}}}^{\textup{teacher}}\left(\textbf{X}_{\textup{1}},\, \textbf{X}_{\textup{2}}\right)\right)
\end{equation}
$\textit{g}$ and $\theta_{\textit{e}}$ correspond to the feature extraction module of the teacher network and its parameters, respectively. $\textit{f}$ and $\theta_{\textit{r}}$ correspond to the information reconstruction module of the teacher network and its parameters, respectively. The loss function to the teacher network is defined as:
\begin{equation}
\textit{L}^{\text {teacher}}=\sum_{\textit{k}=\textup{1}}^{\textit{N}} \sum_{\textit{i}=\textup{1}}^{\textit{H}} \sum_{\textit{j}=\textup{1}}^{\textit{W}} \textup{pc}_{\textup{1}}^{*}(\textit{i},\, \textit{j}) * \textit{l}_{\textup{bce}}\left(\textup{cm}_{\textup{1}}(\textit{i}, \,\textit{j}), \textup{di}_{\textup{2}}(\textit{i}, \,\textit{j}\right))
\end{equation}
where $\textup{pc}_{\textup{1}}^{*}(\textit{i},\, \textit{j})$ is defined as follows:
\begin{equation}
\textup{pc}_{\textup{1}}^{*}(\textit{i},\, \textit{j})=\left\{\begin{array}{ll}
\textup{pc}_{\textup{1}}(\textit{i},\, \textit{j})& \textup{pc}_{\textup{1}}(\textit{i},\, \textit{j}) \geq \alpha \\
\textup{0}& \textup{otherwise}
\end{array}\right.
\end{equation}
$\alpha$ is a threshold we set. We hope the effect of unreliable label pixels on network training can be suppressed. Therefore, $\textup{pc}_{\textup{1}}^{*}(\textit{i},\, \textit{j})$ will be set to 0 if the probability that $\textup{cm}_{\textup{1}}(\textit{i}, \,\textit{j})$ is correct label is below $\alpha$. Change detection is treated as a binary classification problem. Therefore, we use the binary cross-entropy loss function $\textit{l}_{\textup{bce}}$ for the pixels located at $(\textit{i}, \,\textit{j})$ in the $\textit{k}_{\textup{th}}$ pair of images of a training batch with a size of $\textit{N}$. $\textit{l}_{\textup{bce}}$ is defined as follows:
\begin{equation}
\textit{l}_{\textup{bce}}\left(\textit{y}_{\textup{1}}, \textit{y}_{\textup{2}}\right)=-\textit{y}_{\textup{1}} * \log \left(\textit{y}_{\textup{2}}\right)-\left(\textup{1}-\textit{y}_{\textup{1}}\right) * \log \left(\textup{1}-\textit{y}_{\textup{2}}\right)
\end{equation}
when the teacher network is trained well, original bitemporal aerial images are fed into the network. We use a fixed threshold to segment DI generated by the trained teacher network. As $\textup{cm}(\textit{i},\,\textit{j})\in\{\textup{0},\textup{1}\}$, we use the middle value 0.5 as the threshold and the change map $\textbf{CM}_{\textup{2}}=\{\textup{cm}_{\textup{2}}(\textit{i},\,\textit{j}), \textup{1}\leq \textit{i}\leq \textit{H}, \textup{1}\leq \textit{j}\leq \textit{W}\}$ as the pseudo label II can be obtained. We still use the designed filter to deal with the pseudo label $\textbf{CM}_{\textup{2}}$. Then the filtered map II $\textbf{PC}_{\textup{2}}=\{\textup{pc}_{\textup{2}}(\textit{i},\,\textit{j}), \textup{0}\leq \textit{i}\leq \textit{H}, \textup{1}\leq \textit{j}\leq \textit{W}\}$, where $\textup{0}\leq \textup{pc}_{\textup{2}}(\textit{i},\,\textit{j})\leq \textup{1}$, can be obtained. $\textbf{PC}_{\textup{2}}$ will be used to control the usage of pixels in $\textbf{CM}_{\textup{2}}$ in the training process of the student network.

\subsubsection{Student Training}
In the student network training stage, given two optical aerial remote sensing images $\textbf{X}_{\textup{1}}$ and $\textbf{X}_{\textup{2}}$, $\textbf{DI}_{\textit{f}}=\{\textup{di}_{\textup{\textit{f}}}(\textup{i},\,\textup{j}), \textup{1}\leq \textit{i}\leq \textit{H}, \textup{1}\leq \textit{j}\leq \textit{W}\}$ is the output of the student network, which can be defined as follows:
\begin{equation}
\textbf{DI}_{\textit{f}}=\varphi_{\eta_{\textit{r}}}^{\text {student}}\left(\xi_{\eta_{\textit{e}}}^{\text {student}}\left(\textbf{X}\textup{1},\, \textbf{X}\textup{2}\right)\right)
\end{equation}
$\xi$ and $\eta_{\textit{e}}$ correspond to the feature extraction module of the student network and its parameters, respectively. $\varphi$ and $\eta_{\textit{r}}$ correspond to the information reconstruction module of the student network and its parameters, respectively. We use $\textbf{CM}_{\textup{1}}$ and $\textbf{CM}_{\textup{2}}$ to jointly train the student network. The loss function based on $\textbf{CM}_{\textup{1}}$ of the student network is defined as:
\begin{equation}
\textit{L}_{\textup{1}}^{\text {student}}=\sum_{\textit{k}=\textup{1}}^{\textit{N}} \sum_{\textit{i}=\textup{1}}^{\textit{H}} \sum_{\textit{j}=\textup{1}}^{\textit{W}} \textup{pc}_{\textup{1}}^{*}(\textit{i},\, \textit{j}) * \textit{l}_{\textup{bce}}\left(\textup{cm}_{\textup{1}}(\textit{i}, \,\textit{j}), \textup{di}_{\textit{f}}(\textit{i}, \,\textit{j}\right))\end{equation}
The loss function based on $\textbf{CM}_{\textup{2}}$ of the student network is defined as:
\begin{equation}
\textit{L}_{\textup{2}}^{\text {student}}=\sum_{\textit{k}=\textup{1}}^{\textit{N}} \sum_{\textit{i}=\textup{1}}^{\textit{H}} \sum_{\textit{j}=\textup{1}}^{\textit{W}} \textup{pc}_{\textup{2}}^{*}(\textit{i},\, \textit{j}) * \textit{l}_{\textup{bce}}\left(\textup{cm}_{\textup{2}}(\textit{i}, \,\textit{j}), \textup{di}_{\textit{f}}(\textit{i}, \,\textit{j}\right))\end{equation}
where $\textup{pc}_{\textup{2}}^{*}(\textit{i},\, \textit{j})$ is obtained by the same strategy as $\textup{pc}_{\textup{1}}^{*}(\textit{i},\, \textit{j})$:
\begin{equation}
\textup{pc}_{\textup{2}}^{*}(\textit{i},\, \textit{j})=\left\{\begin{array}{ll}
\textup{pc}_{\textup{2}}(\textit{i},\, \textit{j})& \textup{pc}_{\textup{2}}(\textit{i},\, \textit{j}) \geq \alpha \\
\textup{0}& \textup{otherwise}
\end{array}\right.
\end{equation}
The global loss function to the student network is defined as:
\begin{equation}
\textit{L}^{\text {student}}=\beta L_{\textup{1}}^{\text {student}}+(\textup{1}-\beta) \textit{L}_{\textup{2}}^{\text {student}}
\end{equation}
$\textup{0} \leq \beta \leq \textup{1}$ is a constant parameter which is set via experiments. Although $\textbf{CM}_{\textup{2}}$ has a better quality than $\textbf{CM}_{\textup{1}}$ in general, it is generated based on $\textbf{CM}_{\textup{1}}$ and new errors may be produced in the generation process. We use the parameter $\beta$ to balance the effect of $\textbf{CM}_{\textup{1}}$ and $\textbf{CM}_{\textup{2}}$ on the training.

The student network trained by the joint pseudo labels have a better performance than the teacher network. That will be proved in the Section V-B. In the testing stage, only the student network is used to detect the changes between the two aerial images. When the student network is trained well, original bitemporal aerial images are fed into the network. Then we still use a fixed threshold which is set to 0.5 to segment DI and the final change detection result can be obtained. Although labels are needed in the training process of the teacher and student network, they are obtained by the traditional method which do not require any human labeling. Therefore, our proposed method is unsupervised, which is of great significance in the absence of labeling data for aerial remote sensing image change detection tasks.

\begin{figure*}
\centering
\subfigure[]{
\includegraphics[width=5.0cm]{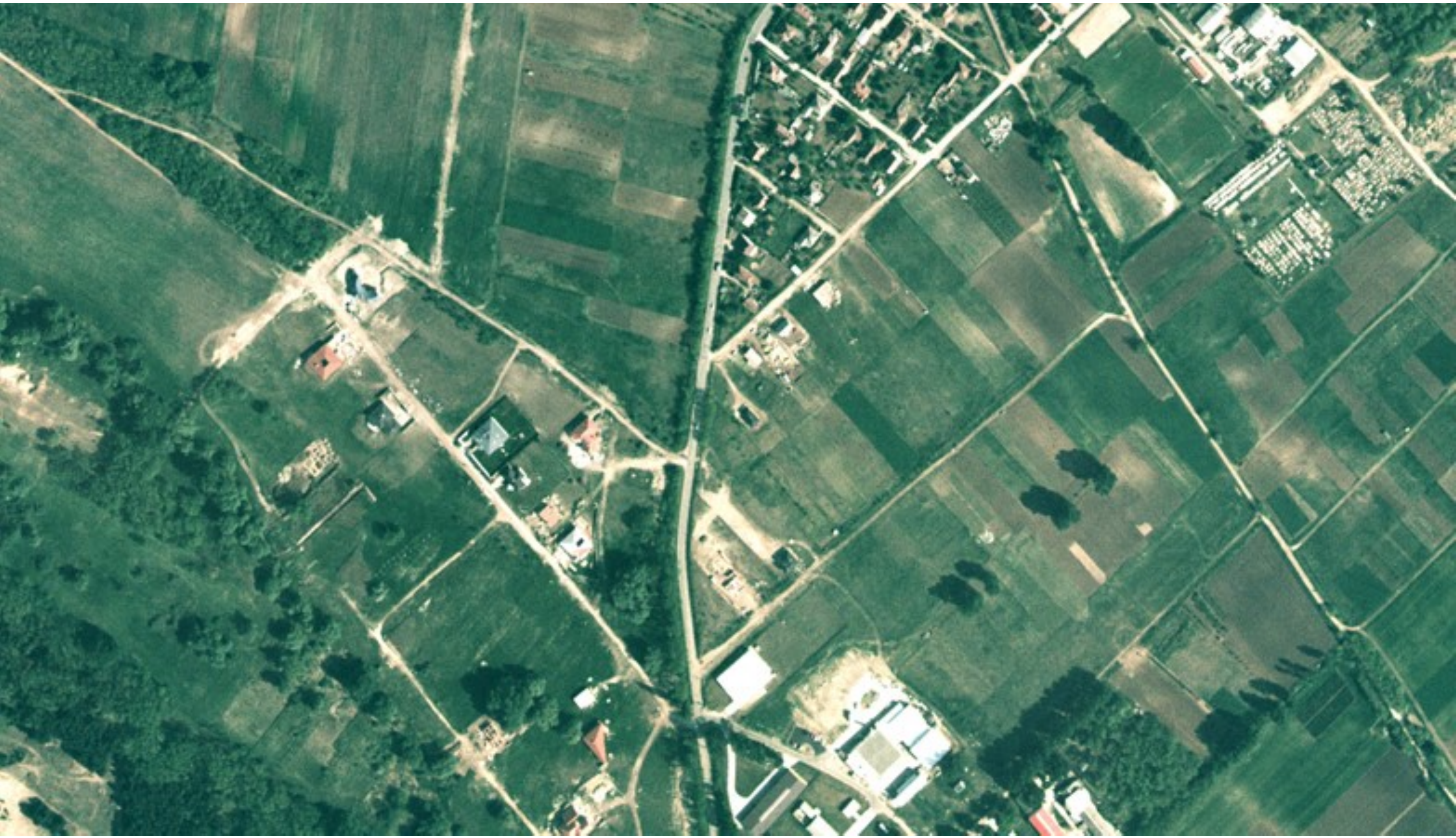}
}
\subfigure[]{
\includegraphics[width=5.0cm]{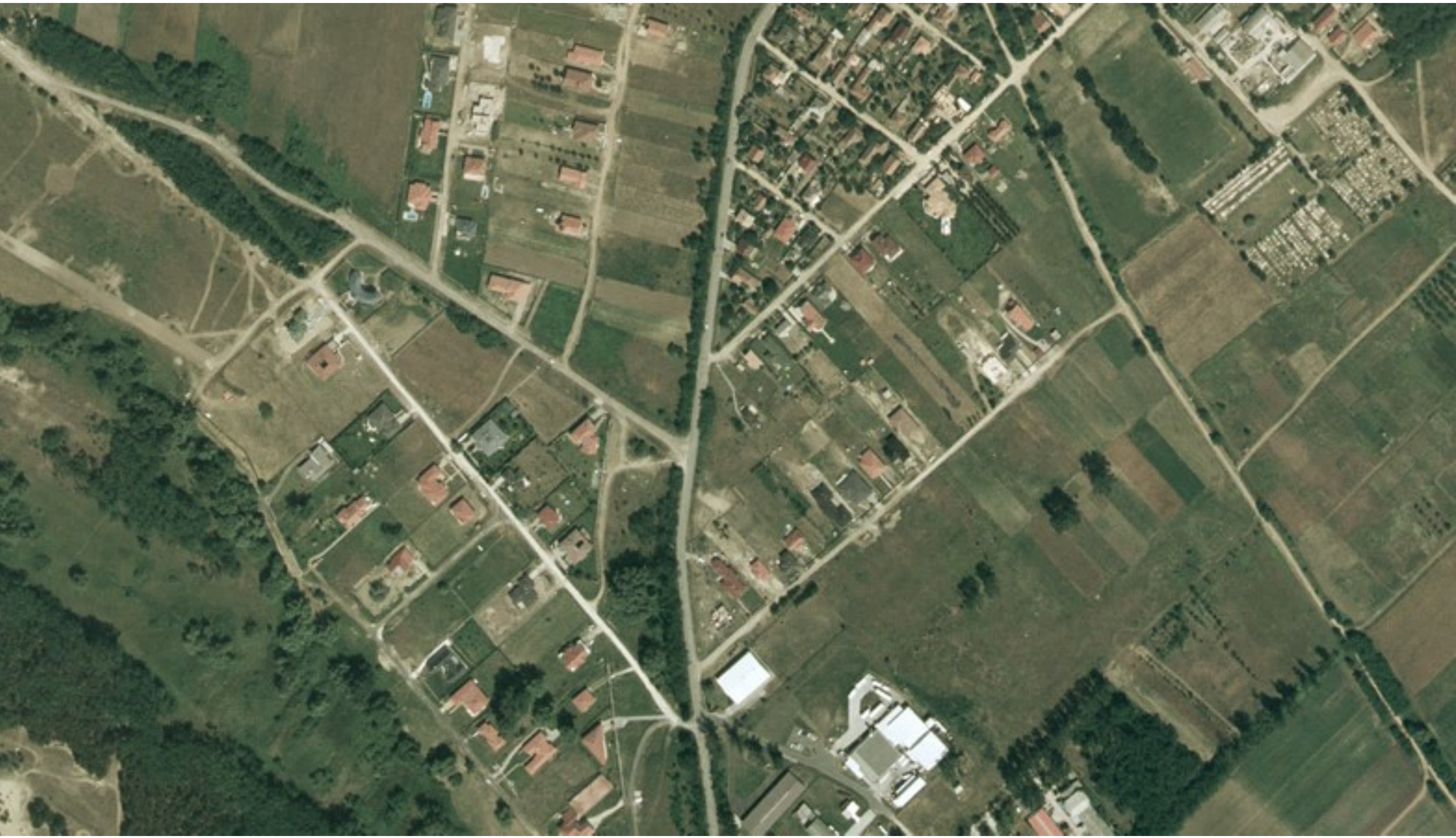}
}
\subfigure[]{
\includegraphics[width=5.0cm]{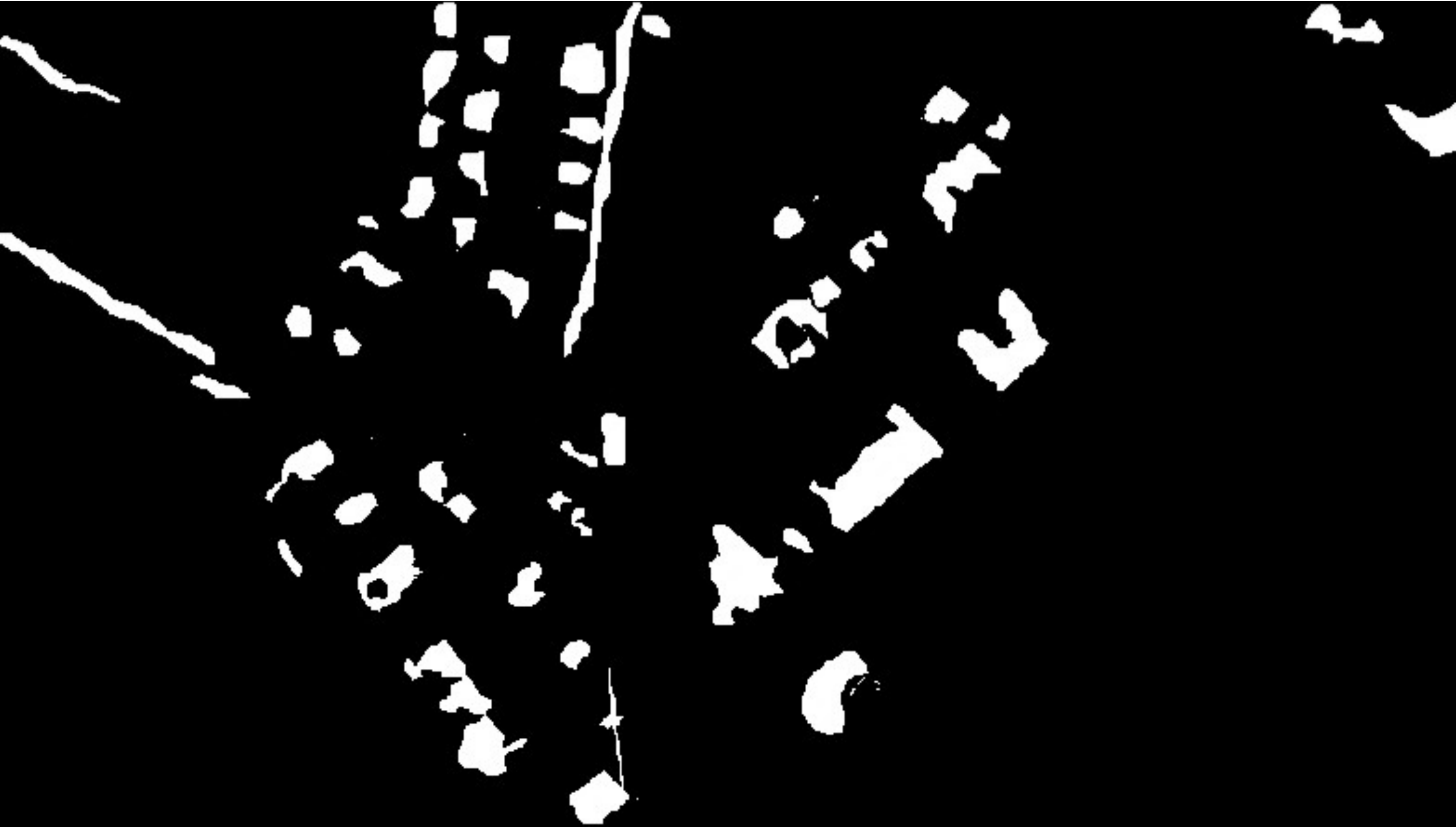}
}
\caption{The SZADA/1 dataset: (a) and (b) are the two images obtained in different time, (c) is the reference change map obtained by manual annotation. Changes are displayed in white regions.}
\label{dataset1}
\end{figure*}

\begin{figure*}
\centering
\subfigure[]{
\includegraphics[width=5.0cm]{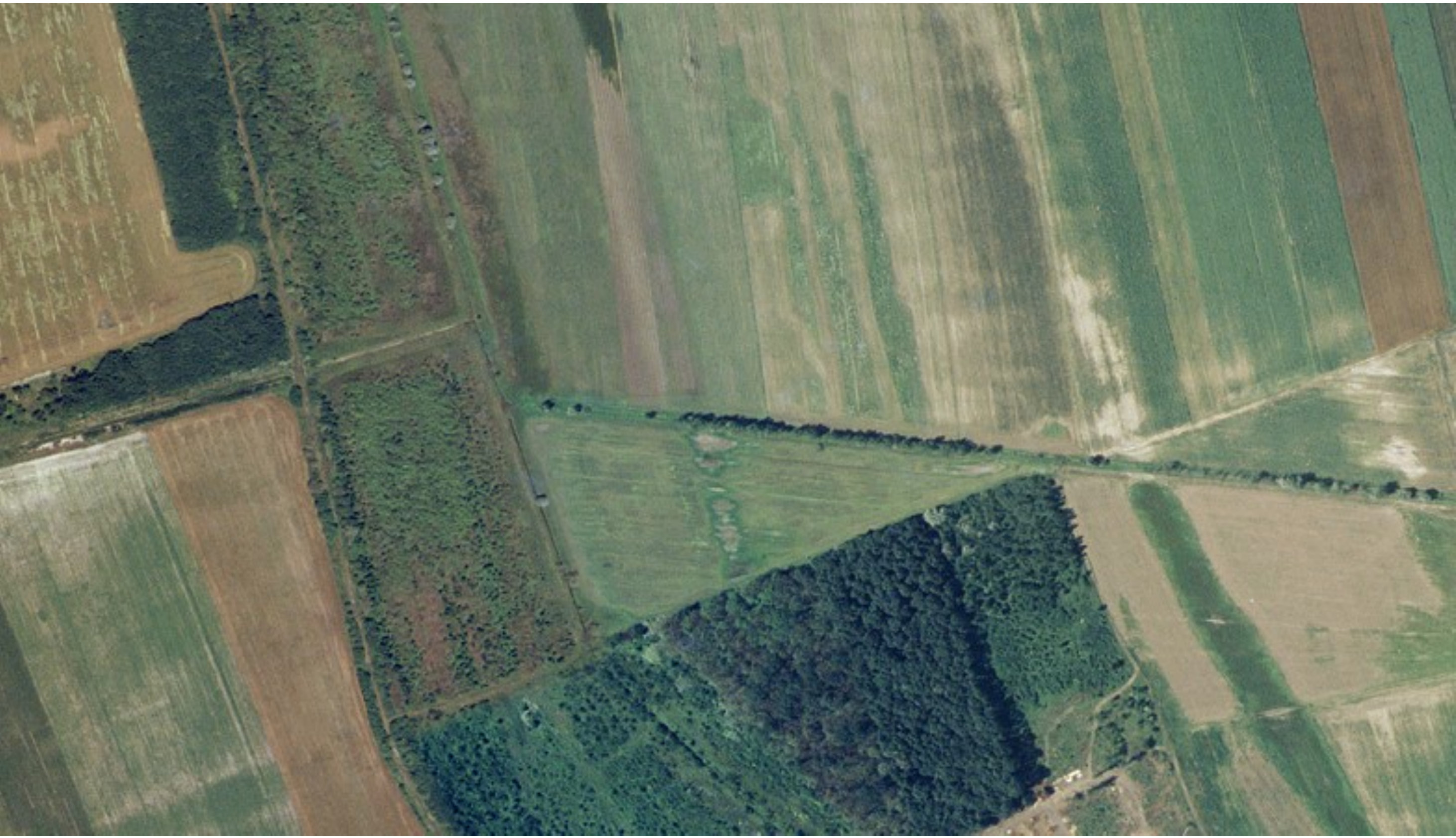}
}
\subfigure[]{
\includegraphics[width=5.0cm]{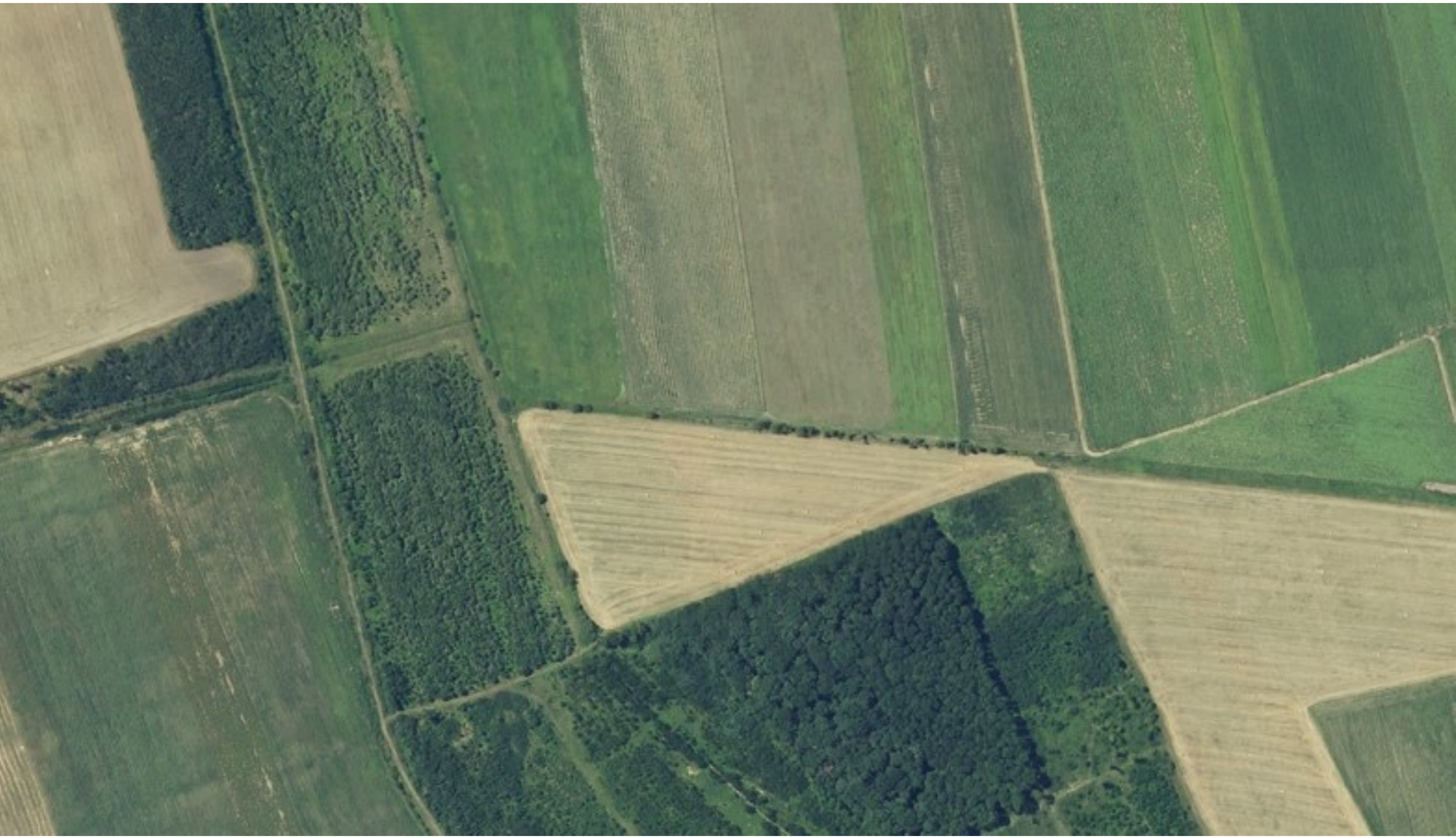}
}
\subfigure[]{
\includegraphics[width=5.0cm]{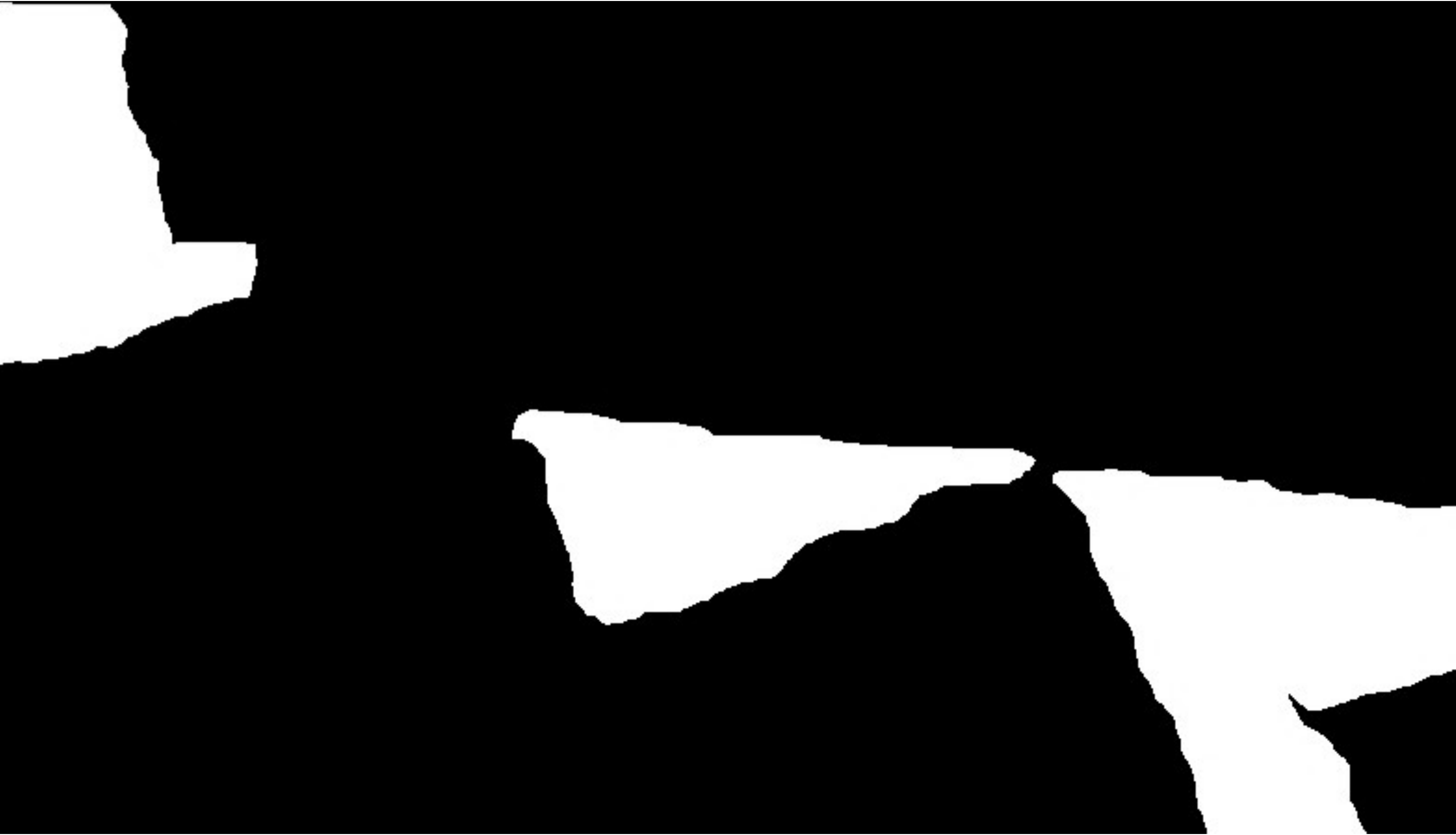}
}
\caption{The TISZADOB/3 dataset: (a) and (b) are the two images obtained in different time, (c) is the reference change map obtained by manual annotation. Changes are displayed in white regions.}
\label{dataset10}
\end{figure*}

\section{Experiments}
\subsection{Datasets}
Our experiments are based on the SZTAKI AirChange Benchmark set which has been widely used in aerial remote sensing image change detection\cite{benedek2009change,singh2014a,zhan2017change,zhang2019triplet-based} and\cite{liu2020convolutional}. This dataset is provided by the Hungarian Institute of Geodesy Cartography and Remote Sensing (F\"{o}MI) and Google Earth\cite{benedek2009change}. The dataset consists of three parts, namely SZADA, TISZADOB, and ARCHIEVE, containing 7, 5, and 1 subdatasets, respectively. Each SZADA dataset contains two aerial remote sensing images which obtained in 2000 and 2005. The size of the images is $\text{952}\times \text{640}$ pixels. Each TISZADOB dataset contains a pair of $\text{952}\times \text{640}$ images obtained in 2000 and 2007, respectively. The ARCHIVE dataset contains a pair of $\text{952}\times \text{640}$ images obtained in 1984 and 2007. The label of each dataset is obtained by expert labeling. The quality of the aerial image taken in 1984 is poor. Therefore, in our experiments, the ARCHIVE dataset is not used. In addition, to meet the requirements of unsupervised tasks, the reference labels in the dataset are only used in the evaluation.

We use the left corner with a size of $\text{784}\times \text{448}$ pixels of aerial images as the test data, which refers the processing in\cite{zhan2017change,zhang2019triplet-based} and\cite{liu2020convolutional}. The rest of the region are cropped to $\text{112}\times \text{112}$ overlappingly to generate training data. To augment training data, we rotate the cropped images by $\text{90}^{\circ}$, $\text{180}^{\circ}$, and $\text{270}^{\circ}$ and flipped them horizontally and vertically. Then we obtain 3744 training image pairs. We evaluate the performance of the proposed method on the sets of images from two datasets, the SZADA/1 and the TISZADOB/3 dataset, as shown in Fig. \ref{dataset1} and Fig. \ref{dataset10}, because the comparison result in most change detection methods on the SZTAKI AirChange Benchmark Set are based on the two set of images.

\subsection{Evaluation Measures and Experimental Settings}
We use three measures to evaluate the performance of the proposed method,  precision (Pr), recall (Rc), and F1-measure (F1), which refers the processing in\cite{zhan2017change,zhang2019triplet-based,8451652} and\cite{liu2020convolutional}.

Precision represents the percentage of the pixels classified correctly into the changed class among all the pixels classified into the changed class. It can be defined as follows:
\begin{equation}
\text{Pr}=\frac{\text{TP}}{\text{TP}+\text{FP}}
\end{equation}
where TP and FP represent the number of the pixels classified into the changed class correctly and the number of the pixels classified into the changed class wrongly, respectively.

Recall represents the percentage of the pixels classified into the changed class correctly among all the actual changed pixels. It can be defined as follows:
\begin{equation}
\text{Rc}=\frac{\text{TP}}{\text{TP}+\text{FN}}
\end{equation}
where FN represents the number of the pixels classified into the unchanged class wrongly.

F-measure is the harmonic mean of precision and recall. It is calculated as follows:
\begin{equation}
\text{F}=\frac{\left(\textit{a}^{\textup{2}}+\textup{1}\right) \cdot \operatorname{\text{Pr}} \cdot \operatorname{\text{Rc}}}{\textit{a}^{\textup{2}} \cdot(\operatorname{\text{Pr}}+\operatorname{\text{Rc}})}
\end{equation}
where $\textit{a}$ is a parameter we set. When $\textit{a}$ is set to 1, F1-measure can be obtained:
\begin{equation}
\text{F} 1=\frac{\textup{2} \cdot \text{Pr} \cdot \text{Rc}}{\text{Pr}+\text{Rc}}
\end{equation}

The teacher and student network are both trained based on the Pytorch on a NVIDIA GeForce GTX TITAN with 12 GB of GPU memory. Adam\cite{sutskever2013on} is used as the optimizer to train the network. We use xavier initializer\cite{glorot2010understanding} to initialize the weights of each layer of the network. The initial learning rate is set to 0.0001.

\begin{figure*}
\centering
\subfigure[]{
\includegraphics[width=4.2cm]{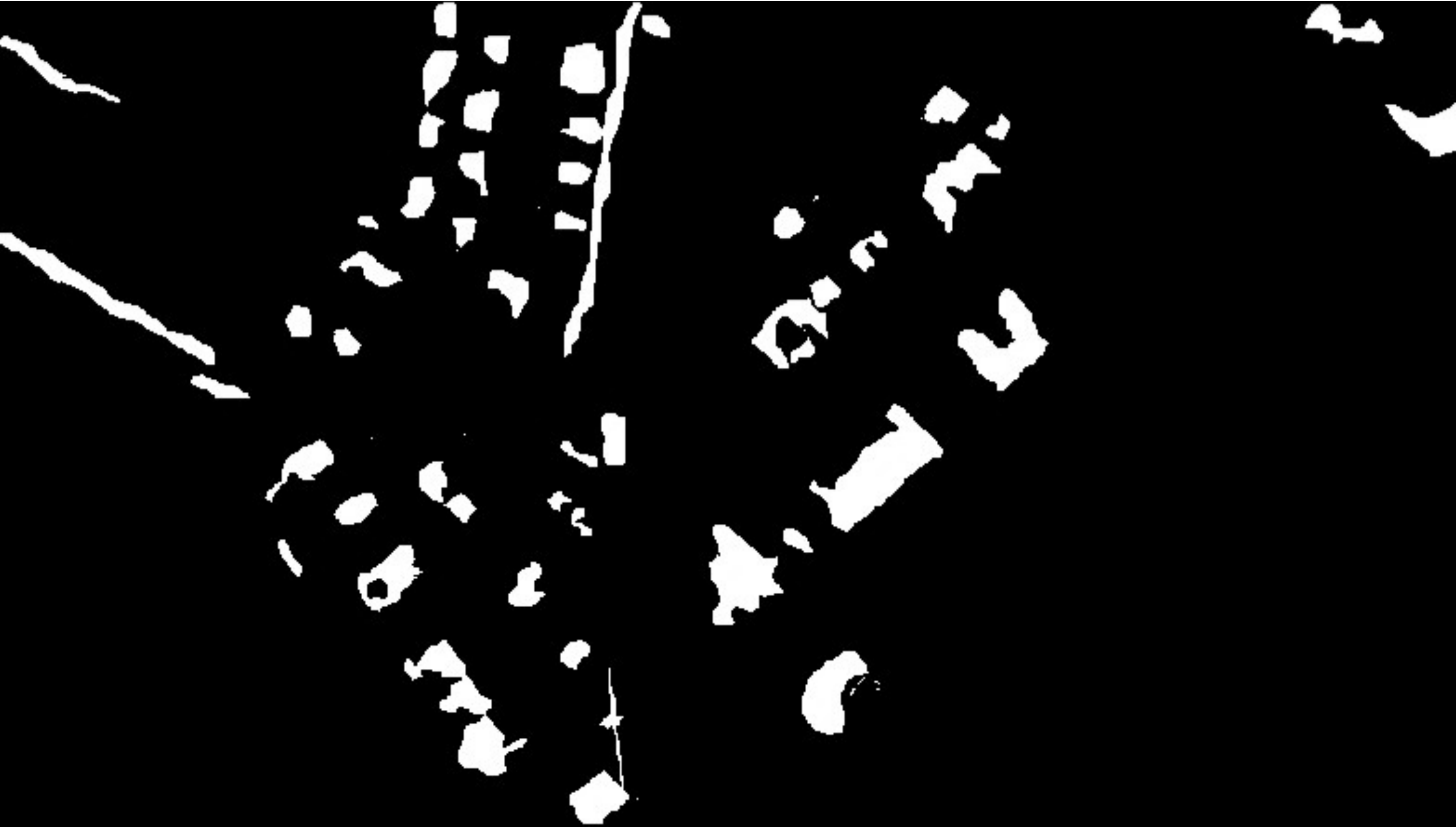}
}
\subfigure[]{
\includegraphics[width=4.2cm]{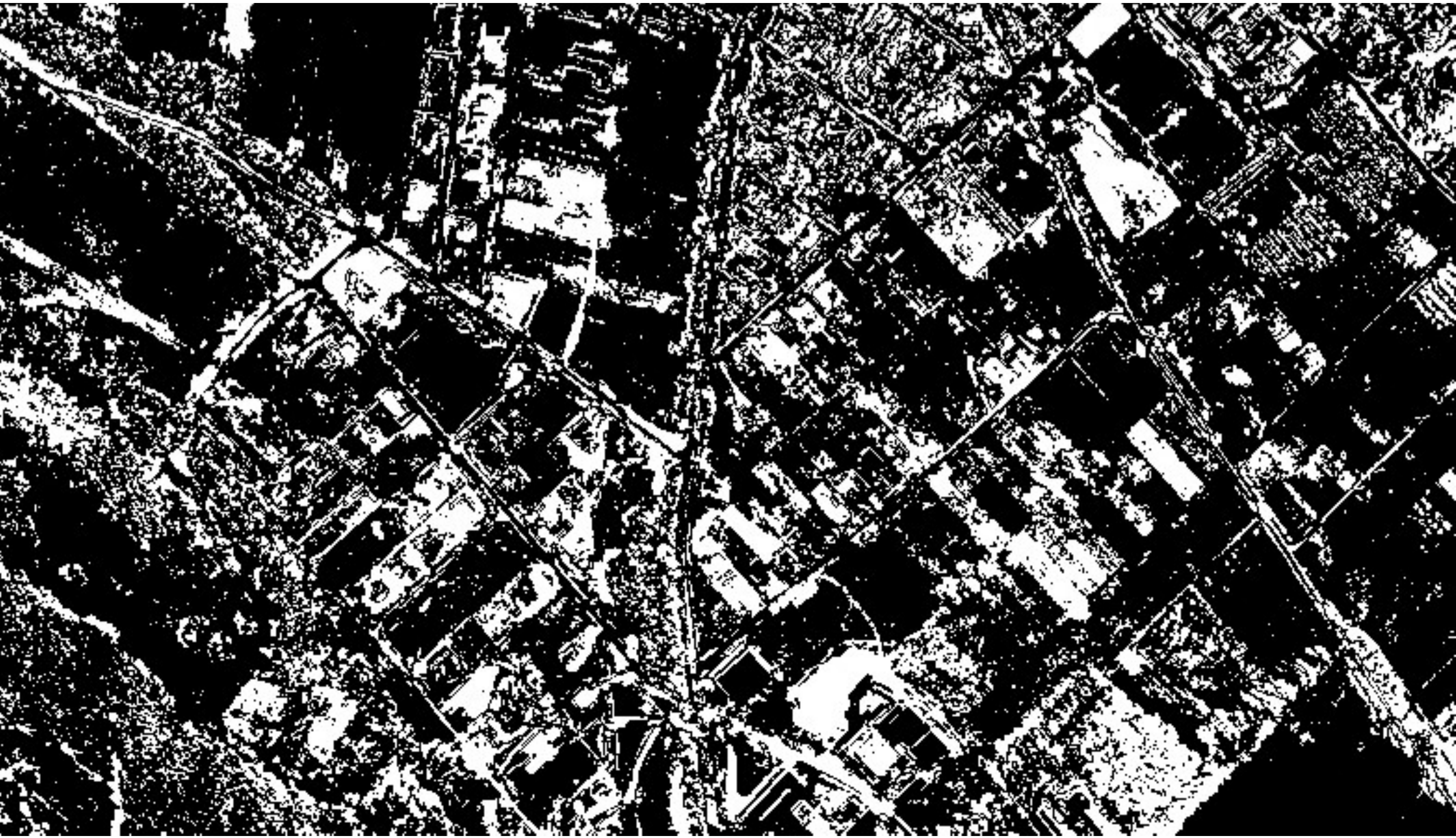}
}
\subfigure[]{
\includegraphics[width=4.2cm]{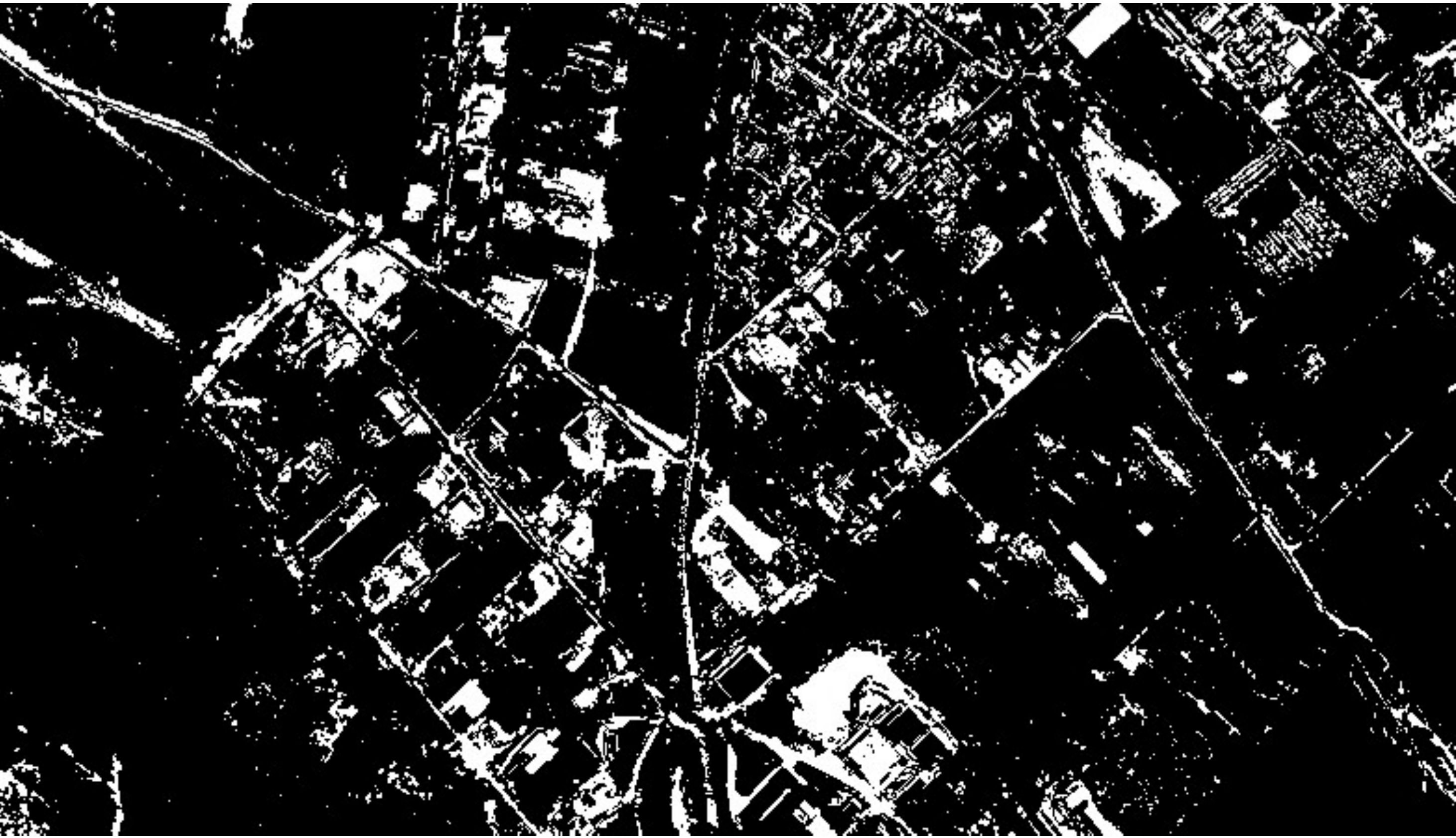}
}
\subfigure[]{
\includegraphics[width=4.2cm]{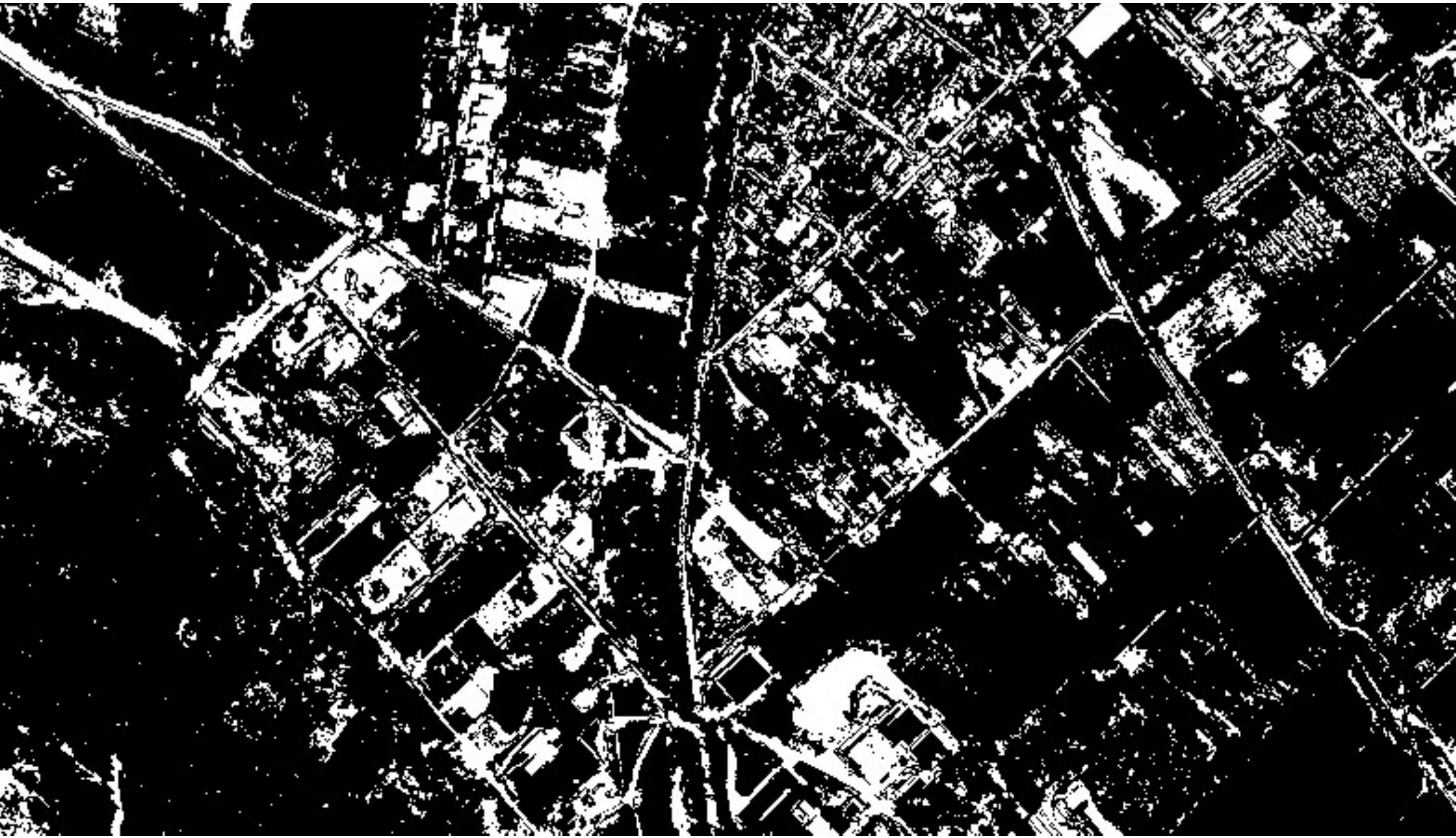}
}
\subfigure[]{
\includegraphics[width=4.2cm]{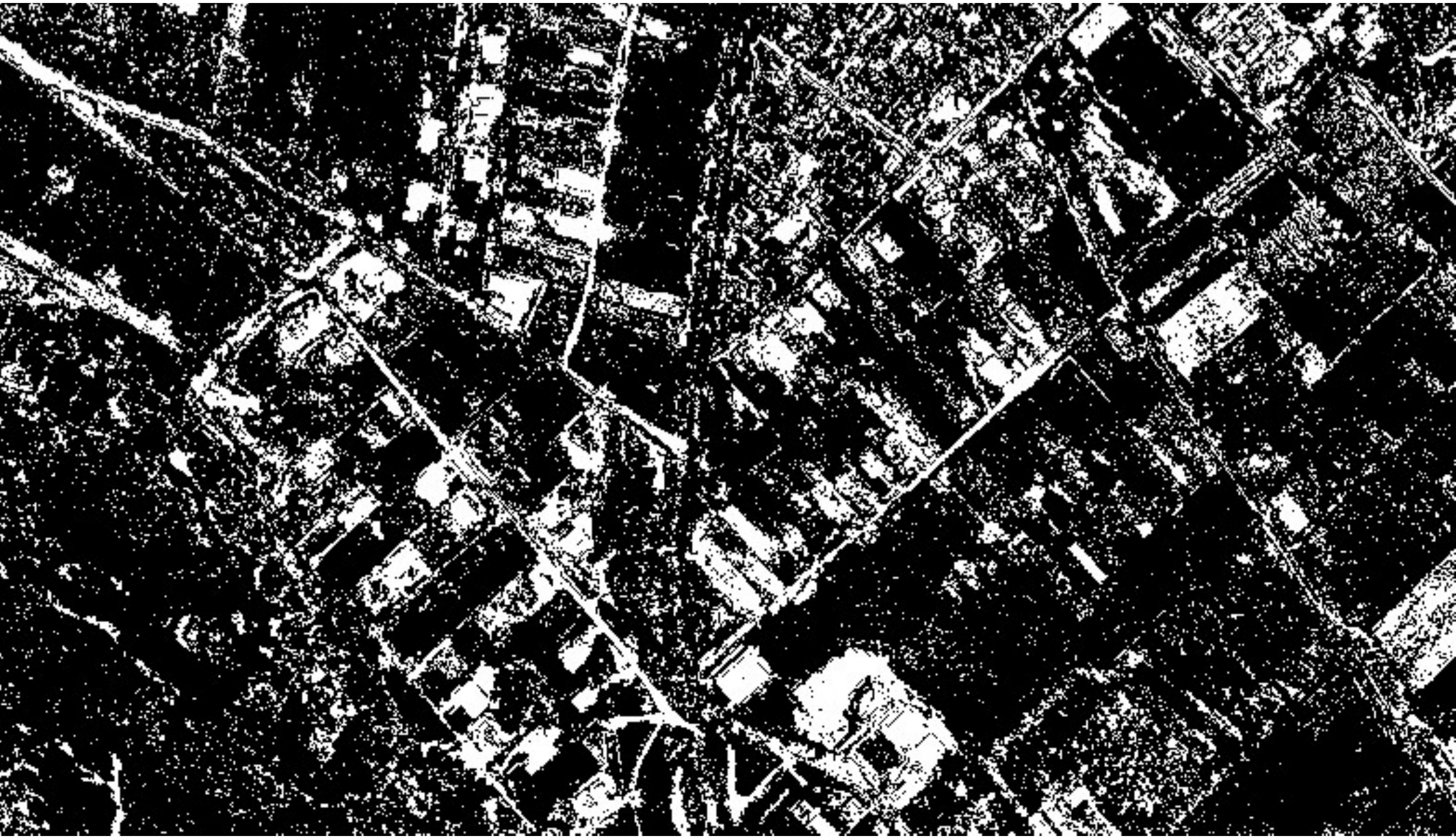}
}
\subfigure[]{
\includegraphics[width=4.2cm]{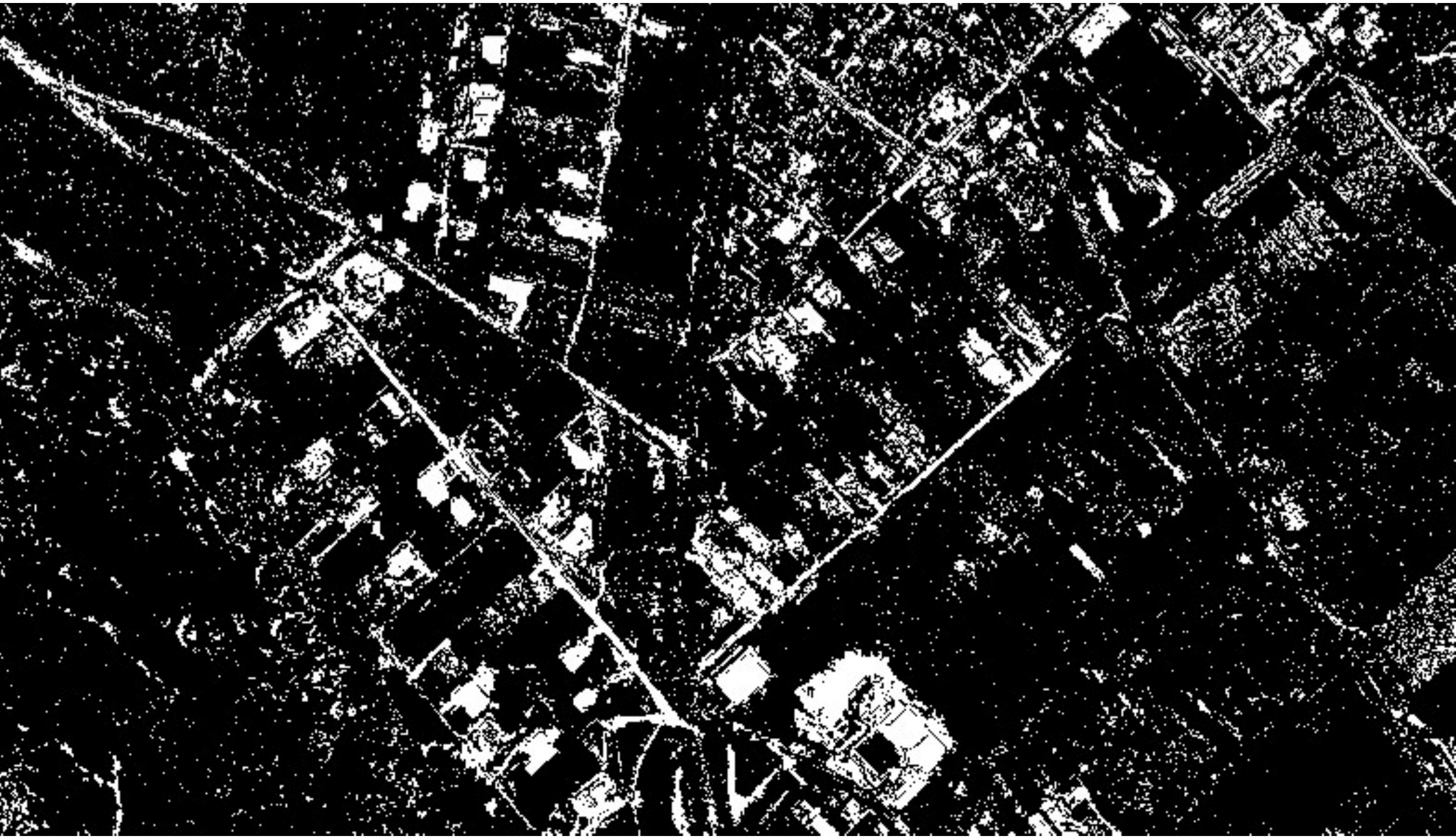}
}
\subfigure[]{
\includegraphics[width=4.2cm]{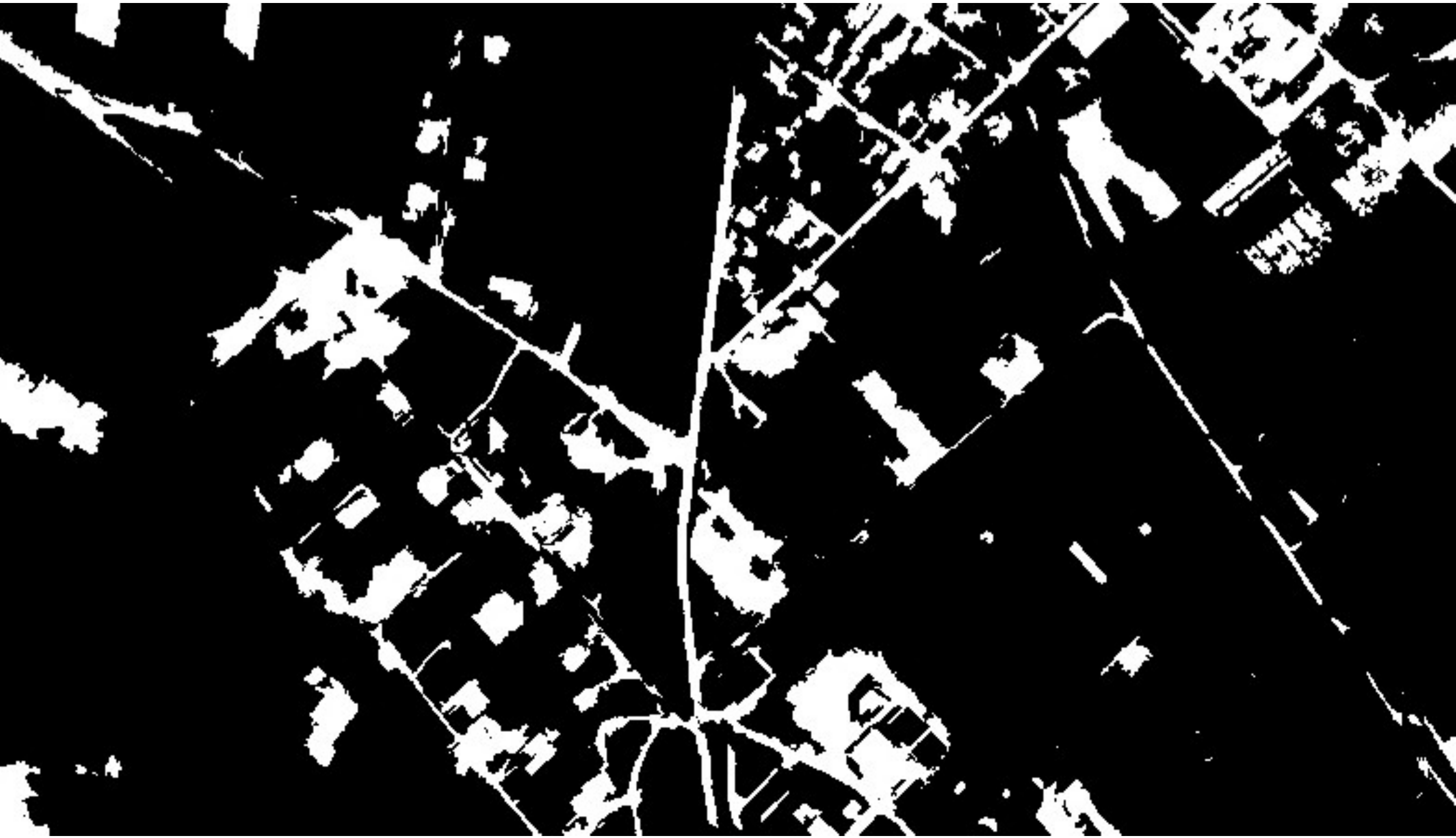}
}
\subfigure[]{
\includegraphics[width=4.2cm]{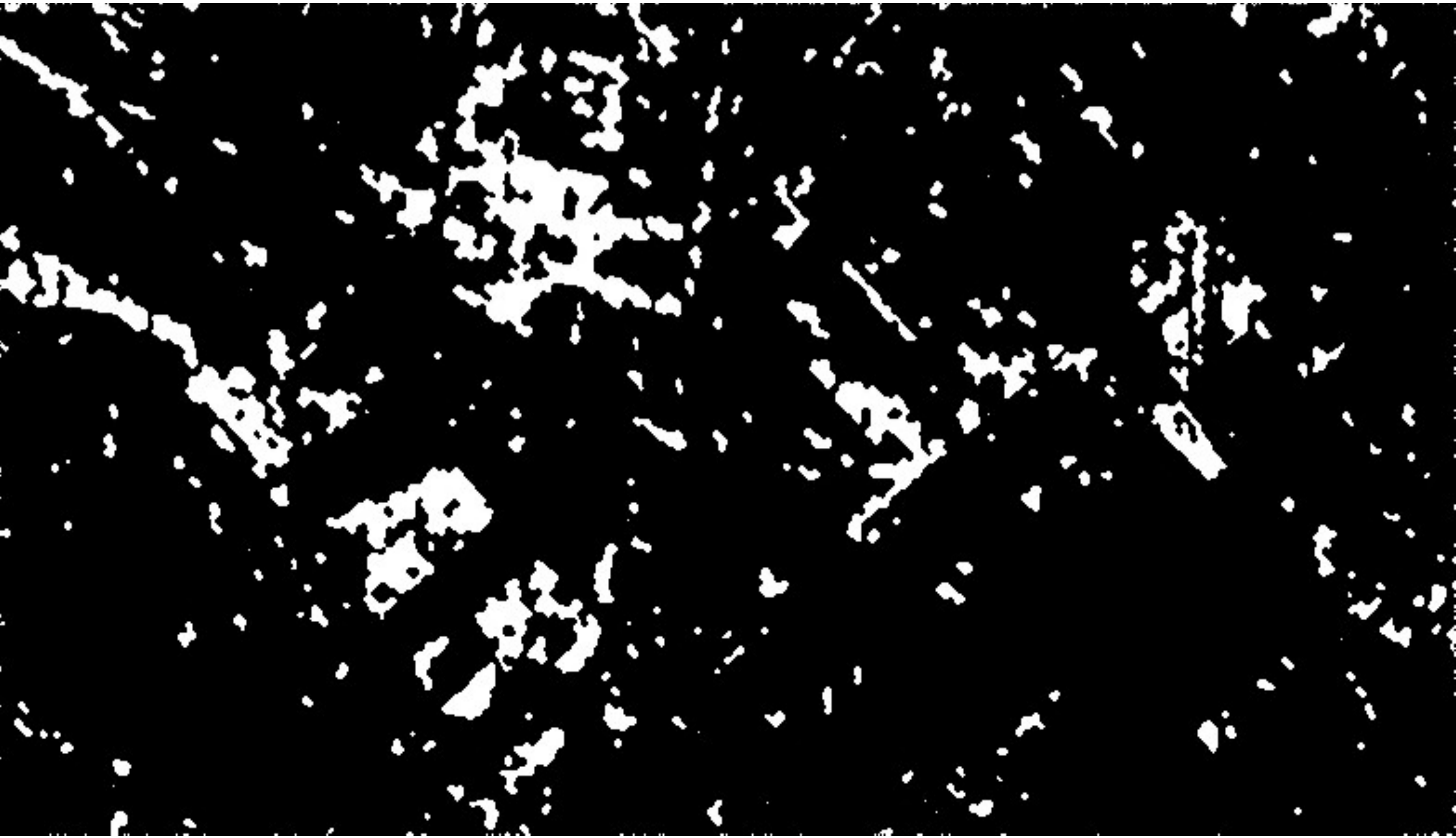}
}
\subfigure[]{
\includegraphics[width=4.2cm]{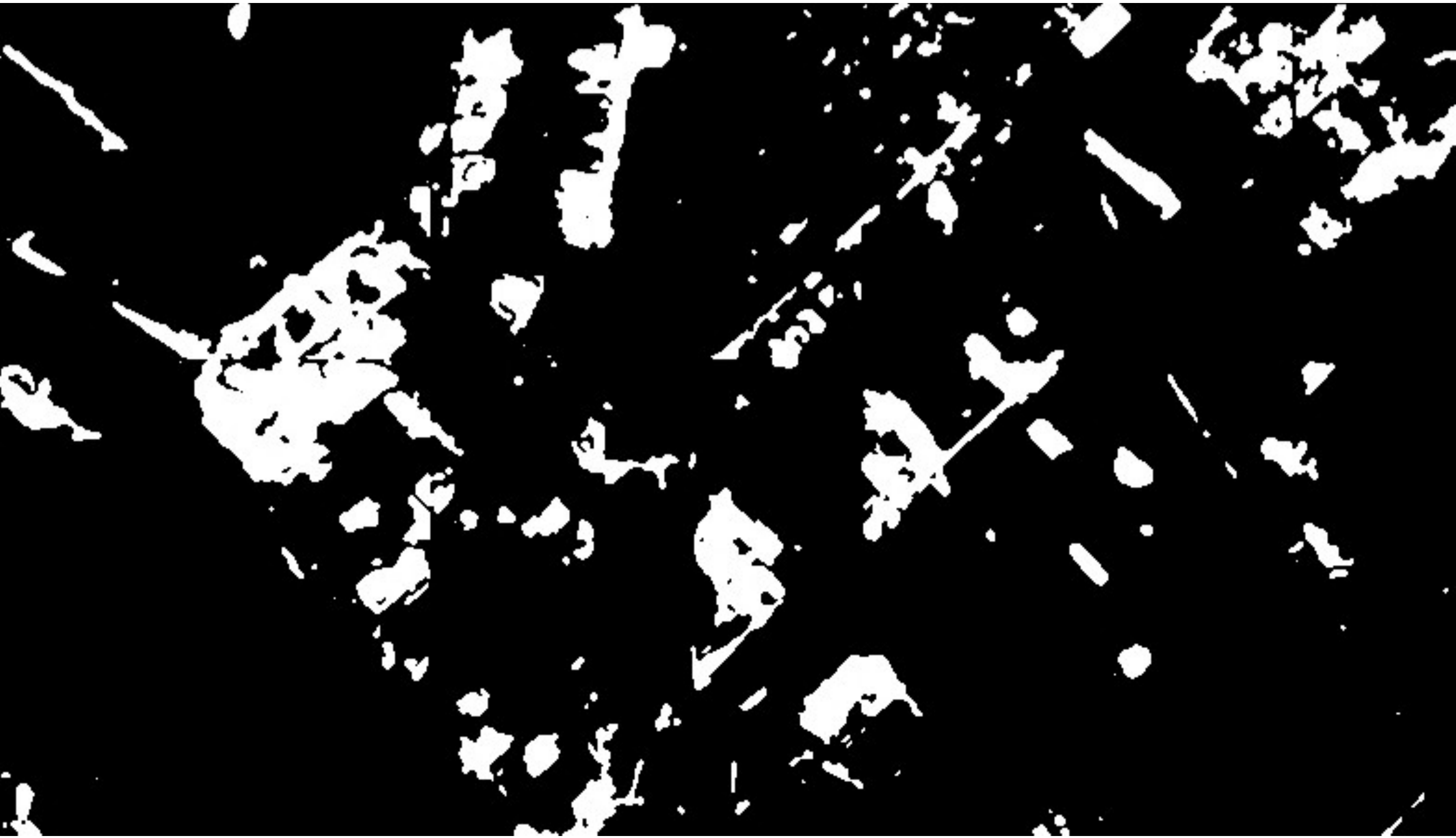}
}
\subfigure[]{
\includegraphics[width=4.2cm]{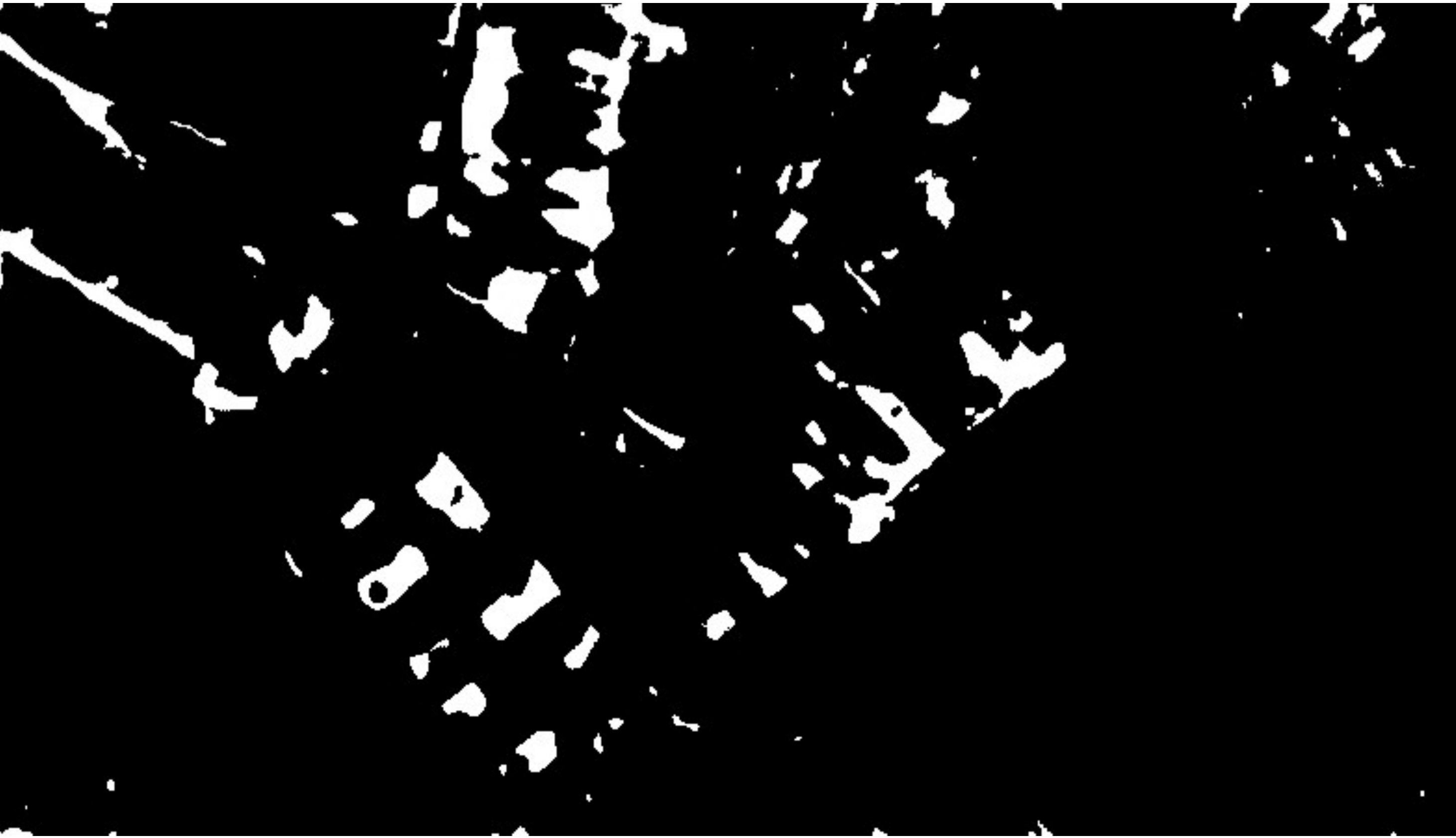}
}
\caption{Change maps obtained by different methods on the SZADA/1 dataset. (a) Reference image. (b) ImageRatio. (c) CVA. (d) PCA. (e) MAD. (f) IRMAD. (g) S3VM. (h) GAN. (i) TLCNN. (j) USTA.}
\label{result1}
\end{figure*}

\begin{table}[]
\centering
\renewcommand{\arraystretch}{1.1}
\caption{Change Detection Result of Each Method on the SZADA/1 Dataset}
\label{result11}
\begin{tabular}{cp{1.5cm}<{\centering}p{1.5cm}<{\centering}p{1.5cm}<{\centering}}
\hline
Method     & Pr(\%)        & Rc(\%)        & F1(\%)    \\ \hline
ImageRatio\cite{howarth1981procedures} & 11.9          & 56.3          & 19.7          \\
CVA\cite{bovolo2007a}        & 20.7          & 47.8          & 28.9          \\
PCA\cite{deng2008pca-based}        & 18.3          & 61.1          & 28.1          \\
MAD\cite{nielsen1998multivariate}        & 17.1          & \textbf{66.8} & 27.2          \\
IRMAD\cite{nielsen2007the}      & 21.5          & 52.3          & 30.5          \\
S3VM\cite{bovolo2008a}       & 19.2          & 48.7          & 27.5          \\
GAN\cite{gong2017generative}        & 21.6          & 34.5          & 26.6          \\
TLCNN\cite{liu2020convolutional}      & 27.2          & 56.1          & 36.7          \\
Ours       & \textbf{47.6} & 44.1          & \textbf{45.8}  \\ \hline
\end{tabular}
\end{table}

\subsection{Results on the SZADA/1 Dataset}
We compare our method with other popular unsupervised change detection methods in the experiment. Fig. \ref{result1} shows the hand-marked label and change maps obtained by the various methods on the SZADA/1 dataset. The dataset contains many small changed areas which make the change detection difficult. Table \ref{result11} lists the quantitative comparison result. The change detection results obtained by traditional methods, including ImageRatio, CVA, PCA, MAD and IRMAD are shown in Fig. \ref{result1}(b) - Fig. \ref{result1}(f). Due to these methods are weak in feature representation, we can find the change maps obtained by them contain lots of white noisy points. In Fig. \ref{result1}(b), ImageRatio yields the most noisy points which lead to the lowest Precision and we can hardly distinguish the main changed areas. In Fig. \ref{result1}(c) and Fig. \ref{result1}(d), the results of CVA and PCA contain less noisy points than ImageRatio at the bottom left and bottom right corner. MAD gains the highest Recall, but Precision and F1-Measure is low due to most pixels classified into changed are wrong noisy points as shown in Fig. \ref{result1}(e). As the extension of MAD, In Fig. \ref{result1}(f), IRMAD generates less noisy points and gains higher F1-Measure than MAD.

The application of training sets has greatly reduced the white noise points in the change map generated by S3VM method, as shown in Fig. \ref{result1}(g). However, many small changed areas are undetected and many unchanged areas are detected as changed, which is particularly serious at the upper right corner of the image. The GAN-based method misdetects lots of small areas, which results in we can barely see the main changed areas as shown in Fig. \ref{result1}(h). The combination of transfer learning and CNNs make the TLCNN method yield a change map with less mistake, but there are still some unchanged areas detected as changed such as the areas at the upper right corner of the image as shown in Fig. \ref{result1}(i). The change map obtained by our proposed method is shown in Fig. \ref{result1}(j). We can see that the problem of misdetected is greatly alleviated. From Table \ref{result11}, we can find that our proposed method have the highest Precision. Besides, Precision and Recall reach a good balance, so the highest F1-Measure is obtained.

\subsection{Results on the TISZADOB/3 Dataset}
\begin{figure*}
\centering
\subfigure[]{
\includegraphics[width=4.2cm]{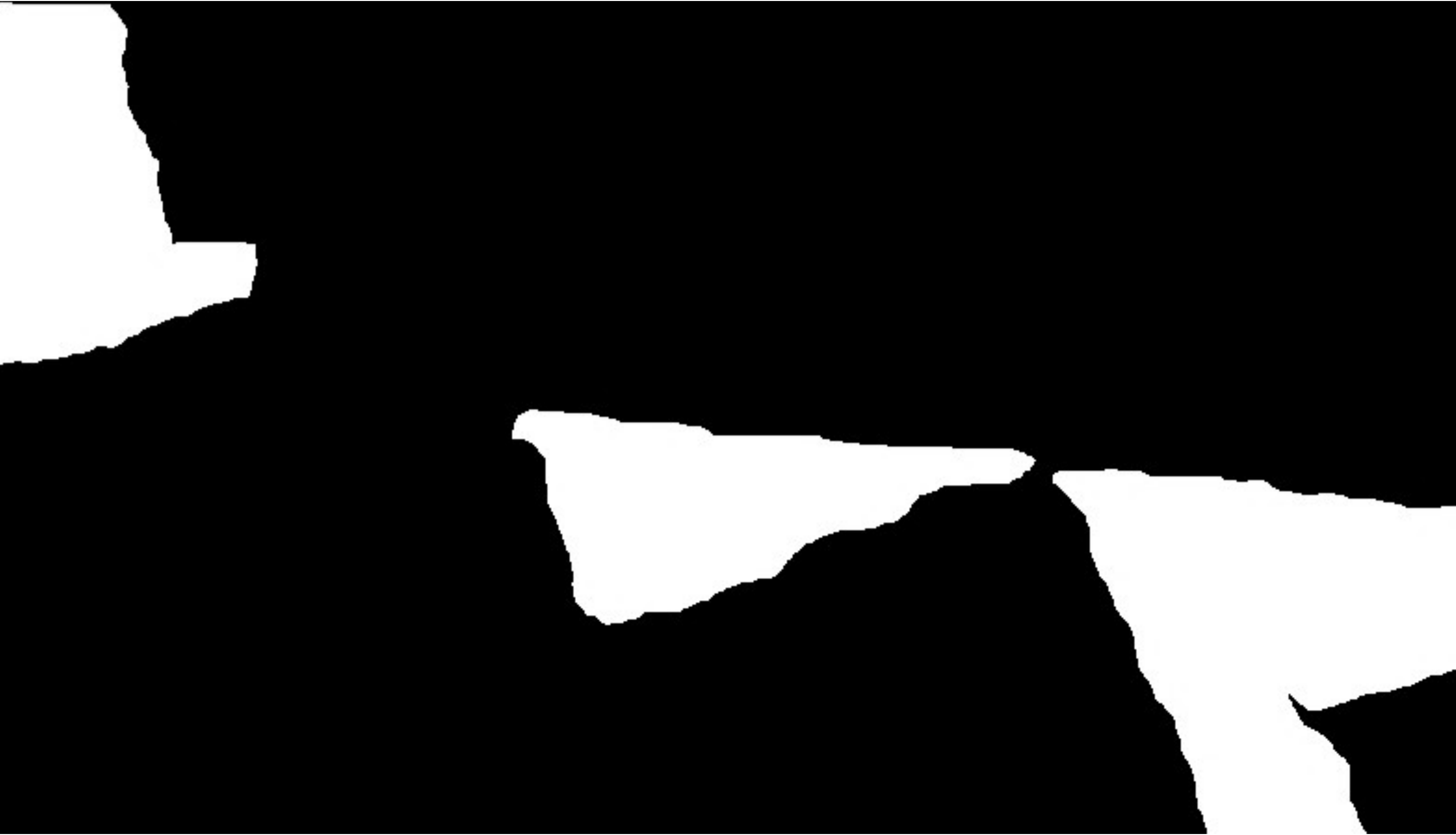}
}
\subfigure[]{
\includegraphics[width=4.2cm]{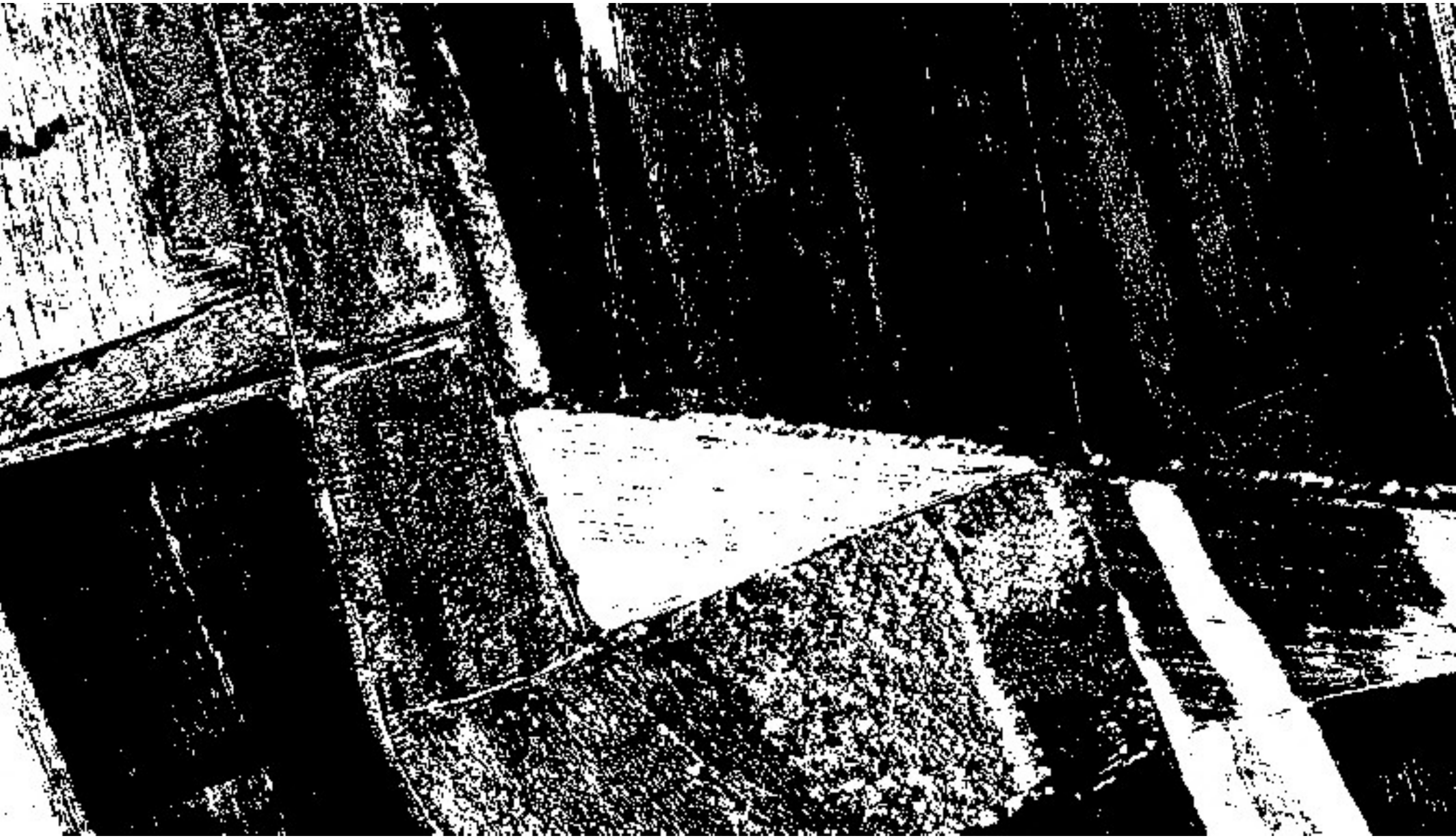}
}
\subfigure[]{
\includegraphics[width=4.2cm]{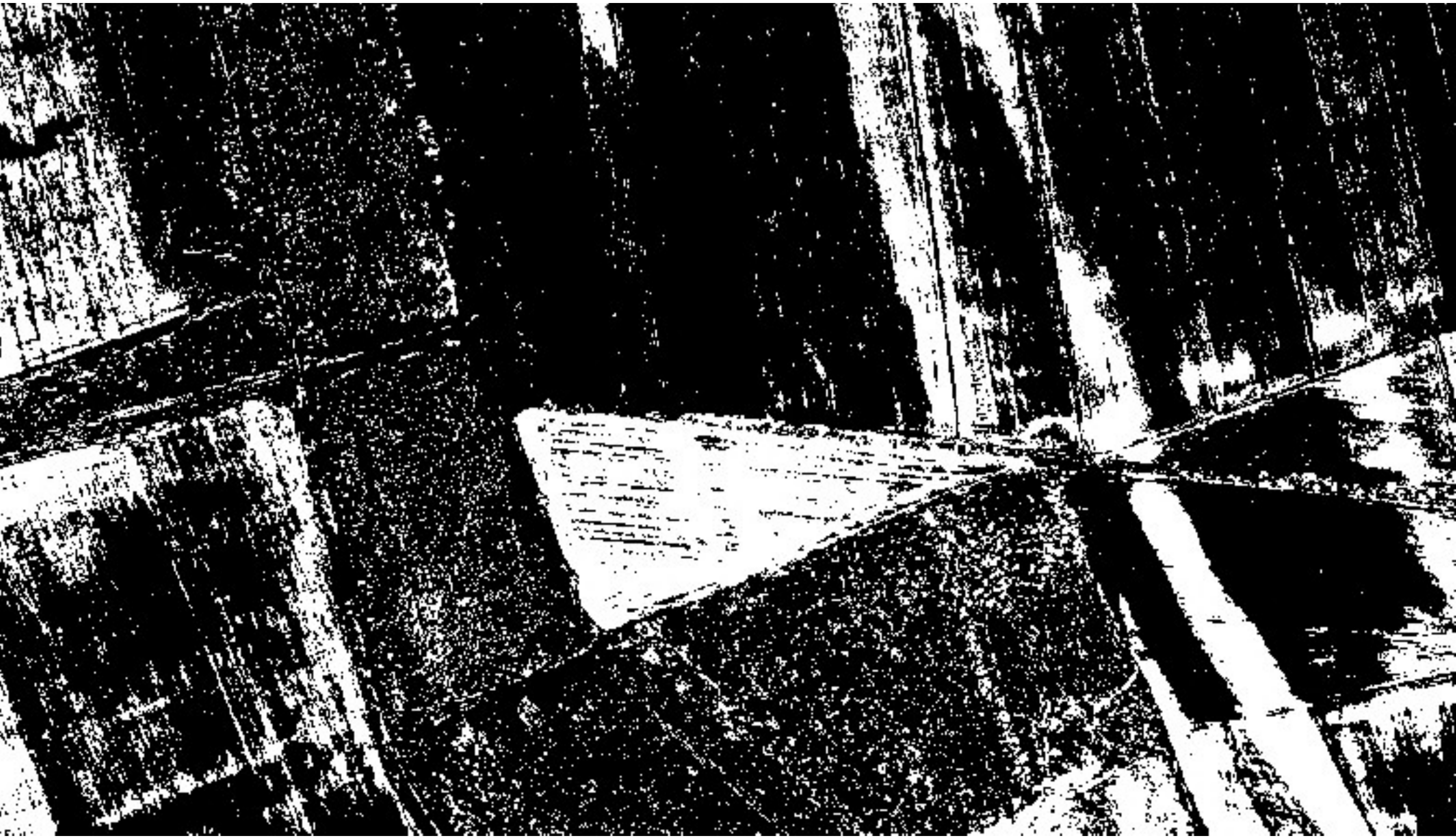}
}
\subfigure[]{
\includegraphics[width=4.2cm]{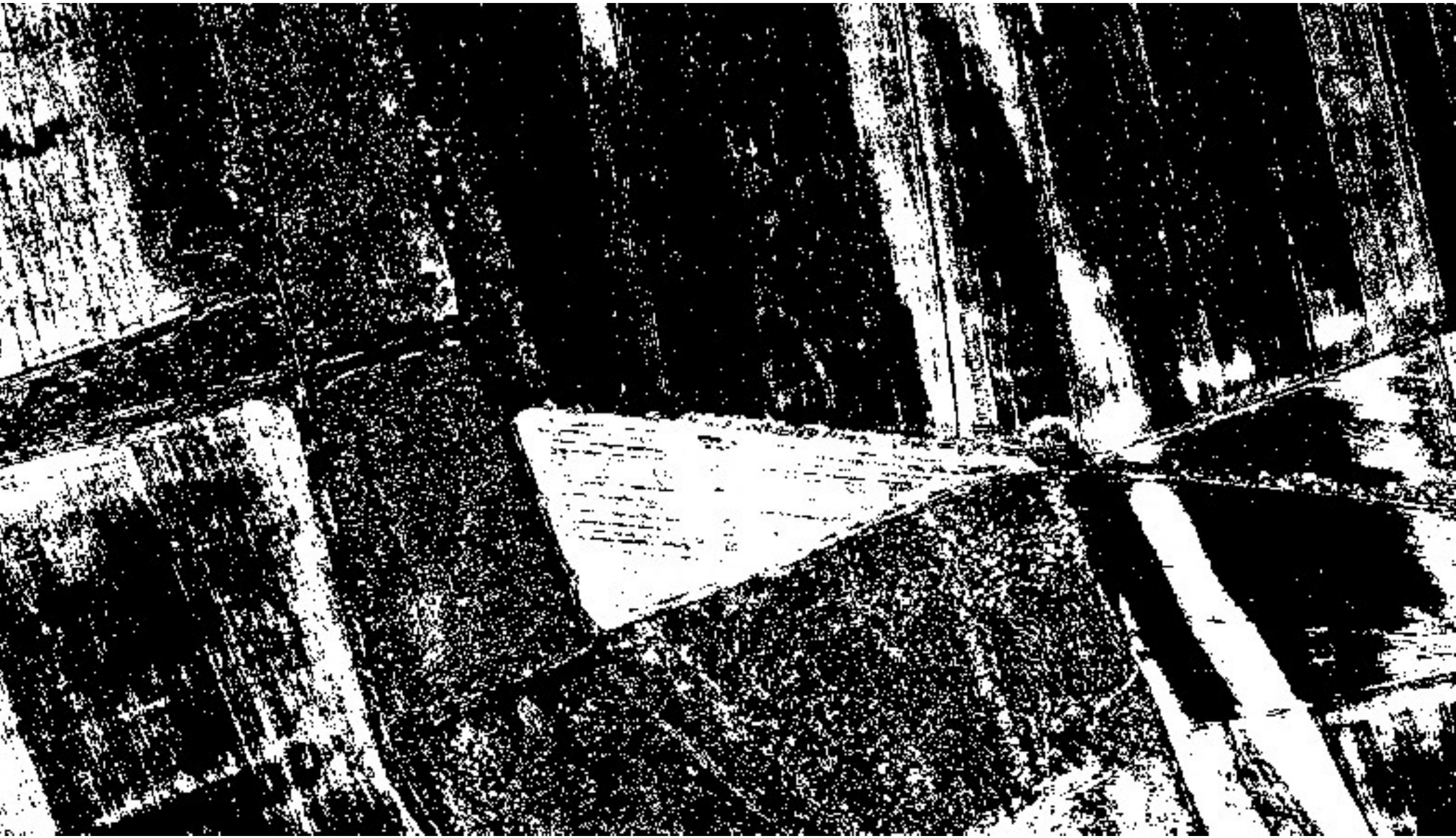}
}
\subfigure[]{
\includegraphics[width=4.2cm]{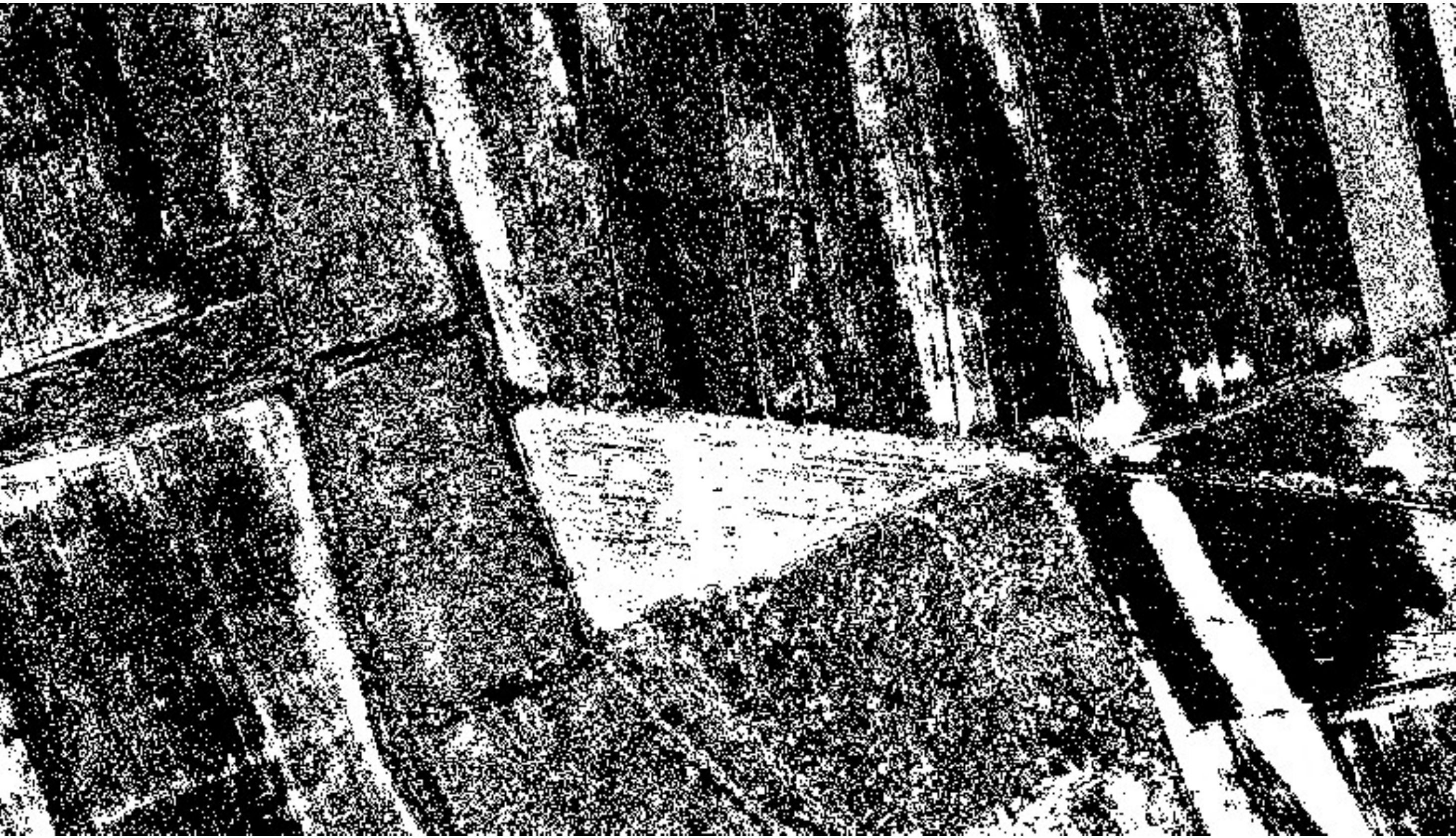}
}
\subfigure[]{
\includegraphics[width=4.2cm]{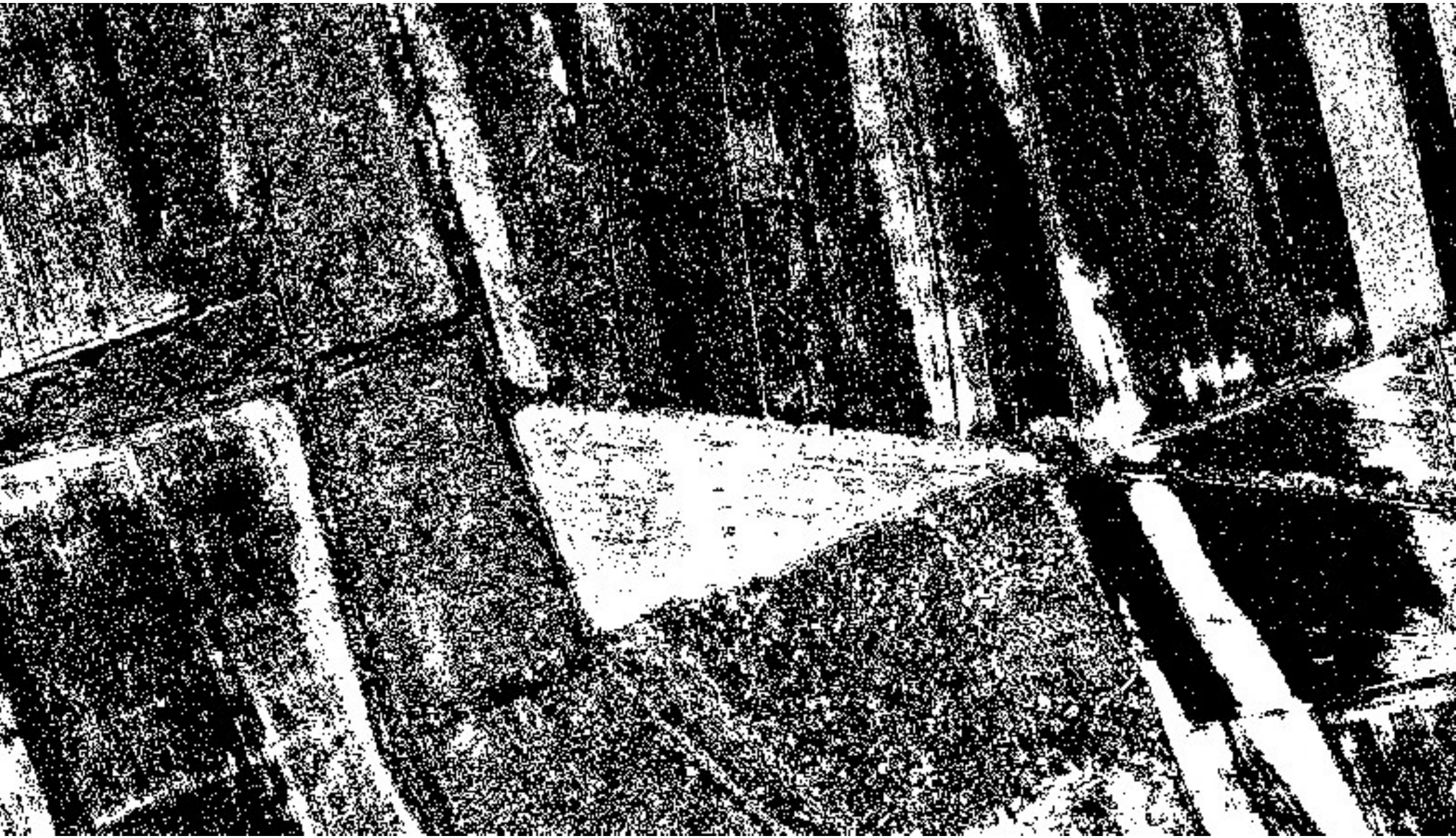}
}
\subfigure[]{
\includegraphics[width=4.2cm]{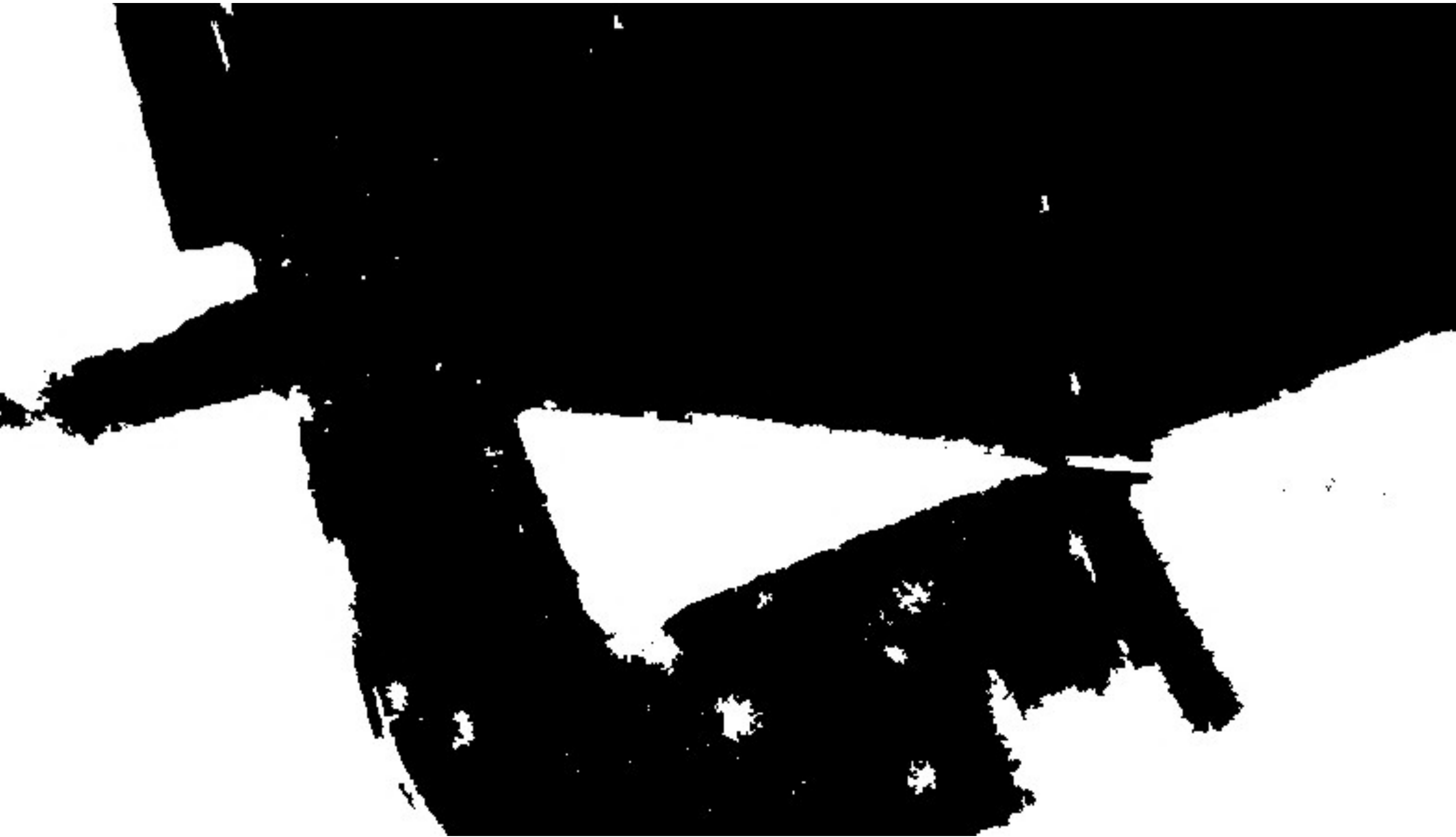}
}
\subfigure[]{
\includegraphics[width=4.2cm]{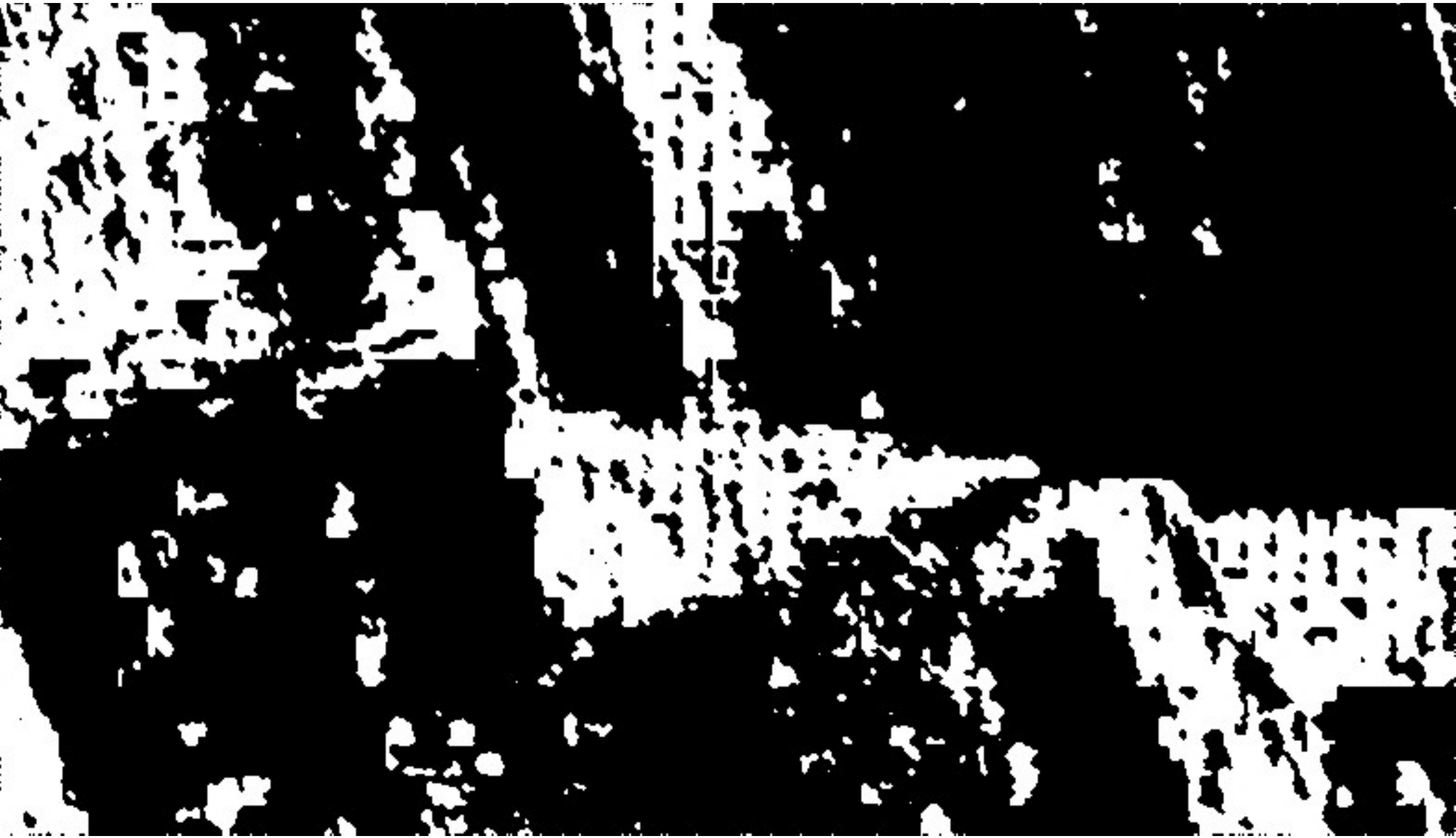}
}
\subfigure[]{
\includegraphics[width=4.2cm]{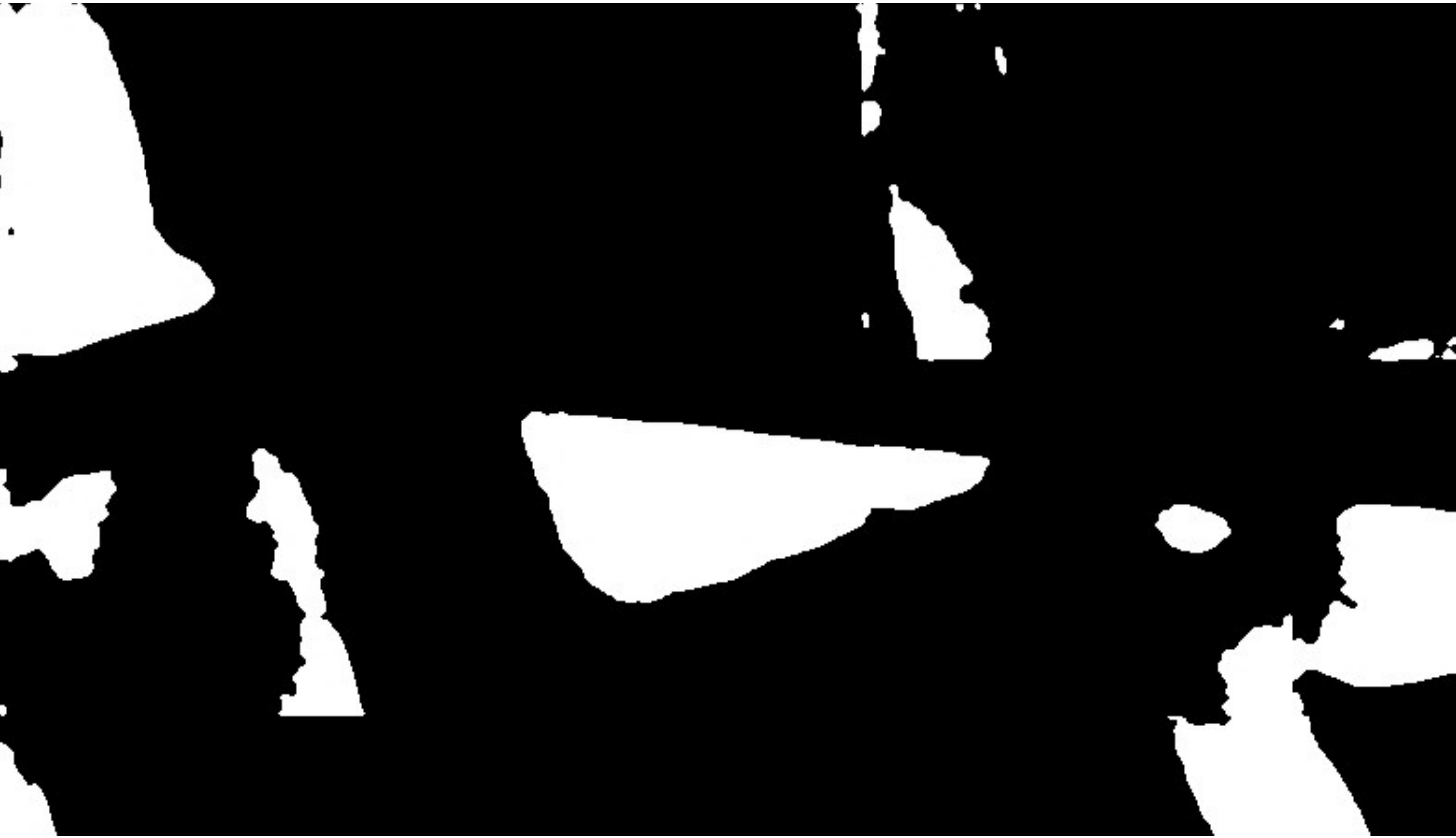}
}
\subfigure[]{
\includegraphics[width=4.2cm]{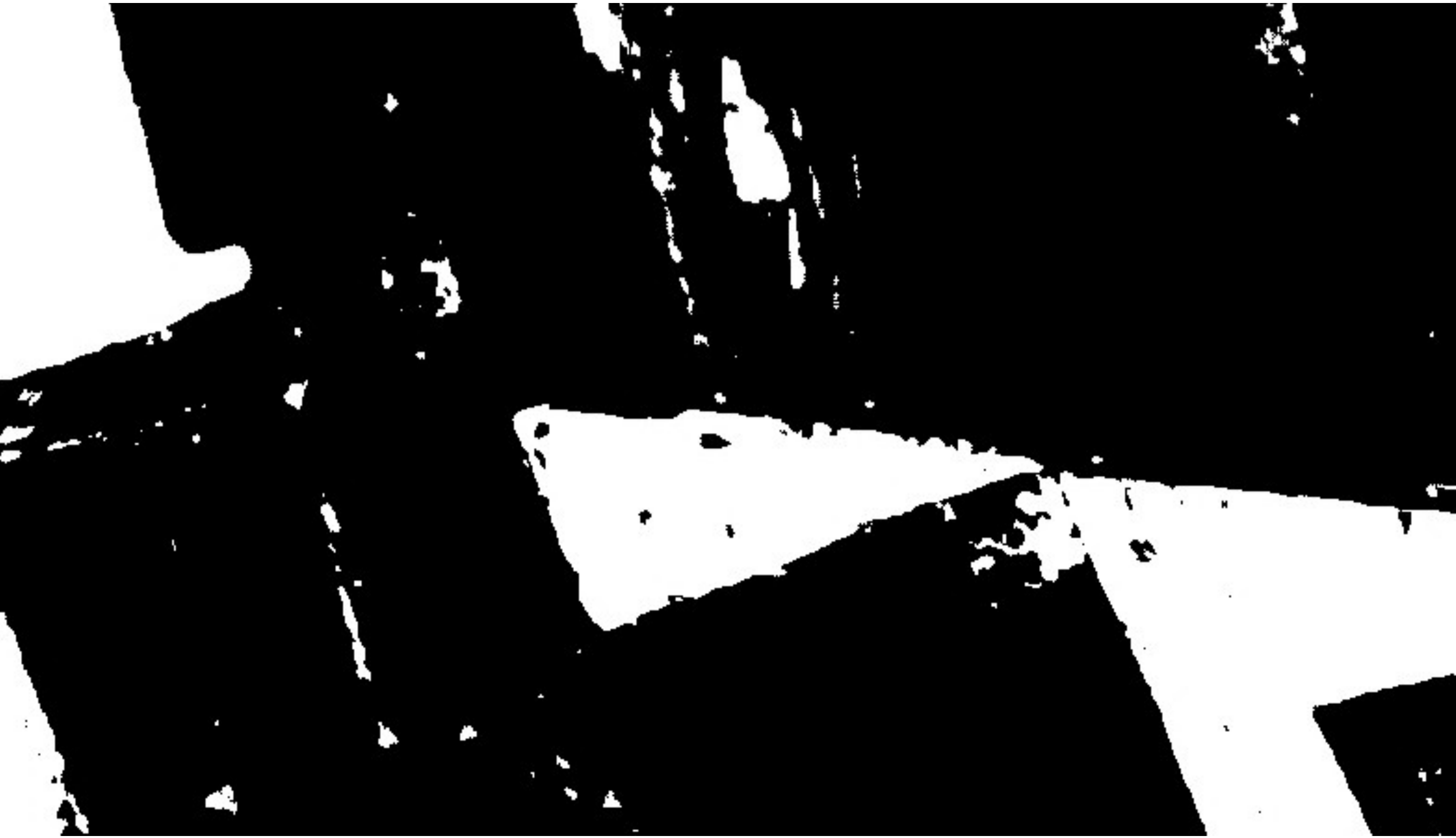}
}
\caption{Change maps obtained by different methods on the TISZADOB/3 dataset. (a) Reference image. (b) ImageRatio. (c) CVA. (d) PCA. (e) MAD. (f) IRMAD. (g) S3VM. (h) GAN. (i) TLCNN. (j) USTA.}
\label{result10}
\end{figure*}

The second experiment is carried on the TISZADOB/3 dataset. We show the hand-marked label and change maps obtained by the various unsupervised methods on the TISZADOB/3 dataset in Fig. \ref{result10}. The quantitative comparison result is listed in Table \ref{result1010}. The results of ImageRatio, CVA, PCA, MAD and IRMAD are shown in Fig. \ref{result10}(b) - Fig. \ref{result10}(f). There are many white noisy points in these change maps. In Fig. \ref{result10}(b), some main changed areas are undetected at the bottom right corner of the image. CVA and PCA detect some unchanged areas as changed at upper right and bottom left corner of the image as shown in Fig. \ref{result10}(c). and Fig. \ref{result10}(d). MAD and IRMAD do not have good performance on this dataset, as we can see that the results of them contain too much noisy points as shown in Fig. \ref{result10}(e) and Fig. \ref{result10}(f).

\begin{table}[]
\centering
\renewcommand{\arraystretch}{1.1}
\caption{Change Detection Result of Each Method on the TISZADOB/3 Dataset}
\label{result1010}
\begin{tabular}{cp{1.5cm}<{\centering}p{1.5cm}<{\centering}p{1.5cm}<{\centering}}
\hline
Method     & Pr(\%)        & Rc(\%)        & F1(\%)    \\ \hline
ImageRatio\cite{howarth1981procedures} & 46.3          & 67.4          & 57.9          \\
CVA\cite{bovolo2007a}        & 37.6          & 54.0          & 44.3          \\
PCA\cite{deng2008pca-based}        & 36.4          & 57.4          & 44.6          \\
MAD\cite{nielsen1998multivariate}        & 22.8          & 47.5          & 30.8          \\
IRMAD\cite{nielsen2007the}      & 25.3          & 50.9          & 33.8          \\
S3VM\cite{bovolo2008a}       & 43.3          & 91.1          & 58.7          \\
GAN\cite{gong2017generative}        & 52.5          & 77.3          & 62.5          \\
TLCNN\cite{liu2020convolutional}      & 78.7          & 71.7          & 75.0          \\
Ours       & \textbf{81.8} & \textbf{96.3} & \textbf{88.4} \\ \hline
\end{tabular}
\end{table}

Compared to the traditional methods mentioned above, the change maps obtained by the training-based methods contain almost no white noisy points as shown in Fig. \ref{result10}(g) - Fig. \ref{result10}(j). From the result of S3VM as shown in Fig. \ref{result10}(g), we can see that most changed areas can be detected with little noisy points, which lead to a high Recall. However, there is a big unchanged area detected as changed at the bottom left corner of the image. The GAN-based method misdetects many small areas as shown in Fig. \ref{result10}(h). The problem of misdetected is alleviated by TLCNN based transfer learning as shown in Fig. \ref{result10}(i). The change map obtained by our proposed method is shown in Fig. \ref{result10}(j). We can find that almost all the changed areas can be detected. Besides, the areas which is detected as changed mistakenly are much less in this change map. From Table \ref{result1010}, we can that our method is the best with the evaluation of all the three measures. It is obvious that our proposed method have a better performance than other methods in the qualitative and quantitative analysis.

\section{Discussion}

In the learning process of the network, the related factors influence the performance of the algorithm. We will analyze the effects of these factors on the performance of the proposed method in the following.

\subsection{Effects of the Neighborhood Size}

In the process of image filtering, the neighborhood size $\textit{w}$ determines the area that the filter focuses on. If $\textit{w}$ is set to be smaller,  the pixels in the small noise areas may be considered as correctly labels pixels. If $\textit{w}$ is set to be larger, the neighborhood of the pixel contains redundant information and the neighborhood filtering of the correct label pixels may be affected by the surrounding noise pixels. In the experiment, we set $\textit{w}$ to be 3, 5, 7 and 9 for researching the influence of the neighborhood size on the change detection result. The relationship between $\textit{w}$ and the performance of the proposed method is shown in Table \ref{filter size}. When $\textit{w}$ is set to be 3, Recall and Precision are very high on SZADA/1 and TISZADOB/3 dataset, respectively, but the other measures are not good. F1-Measure as a harmonic measure is high on the two dataset when $\textit{w}$ is set to be 5. Besides, Precision is very high on SZADA/1 dataset. When $\textit{w}$ is set to be 7, only Recall is high on TISZADOB/3 dataset. The performance of the proposed method is worse when $\textit{w}$ is set to be 9, because the neighborhood contains too much redundant information. Therefore, it is appropriate that $\textit{w}$ is set to be 5 through the above analysis.
\begin{table}[]
\centering
\renewcommand{\arraystretch}{1.1}
\caption{Change Detection Comparison With Different Filter Size}
\label{filter size}
\begin{tabular}{cccccc}
\hline
\multirow{2}{*}{Dataset}    & \multirow{2}{*}{Metrics} & \multicolumn{4}{c}{Size of filter}                       \\
                            &                          & 3              & 5              & 7              & 9     \\ \hline
\multirow{3}{*}{S1}    & Pr(\%)                   & 27.34          & \textbf{47.10} & 41.91          & 22.94 \\
                            & Rc(\%)                   & \textbf{41.59} & 31.93          & 27.97          & 36.85 \\
                            & F1(\%)                   & 32.99          & \textbf{38.06} & 33.55          & 28.27 \\ \hline
\multirow{3}{*}{T3} & Pr(\%)                   & \textbf{95.09} & 89.60          & 86.57          & 94.19 \\
                            & Rc(\%)                   & 71.78          & 80.38          & \textbf{82.51} & 76.55 \\
                            & F1(\%)                   & 81.81          & \textbf{84.74} & 84.49          & 84.46 \\ \hline
\end{tabular}
\end{table}

\begin{figure}
\centering
\subfigure[]{
\includegraphics[width=4.1cm]{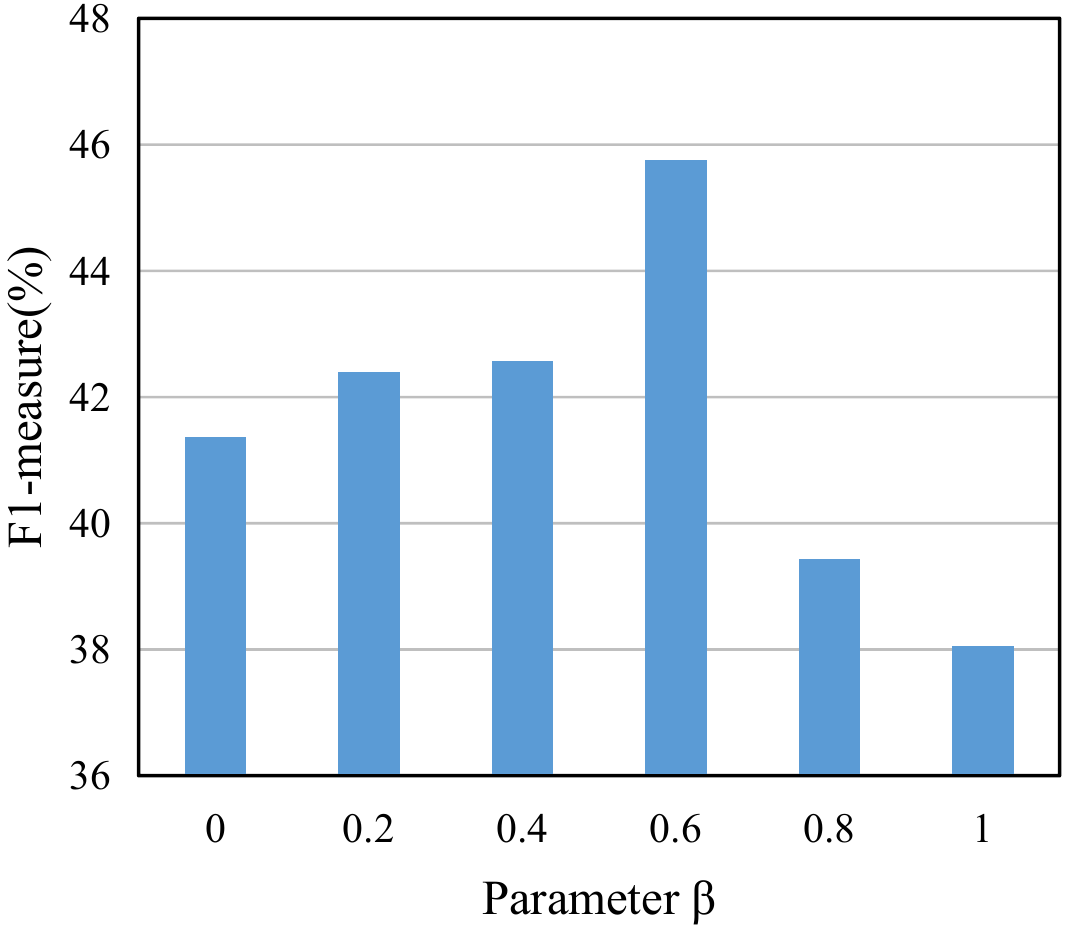}
}
\subfigure[]{
\includegraphics[width=4.1cm]{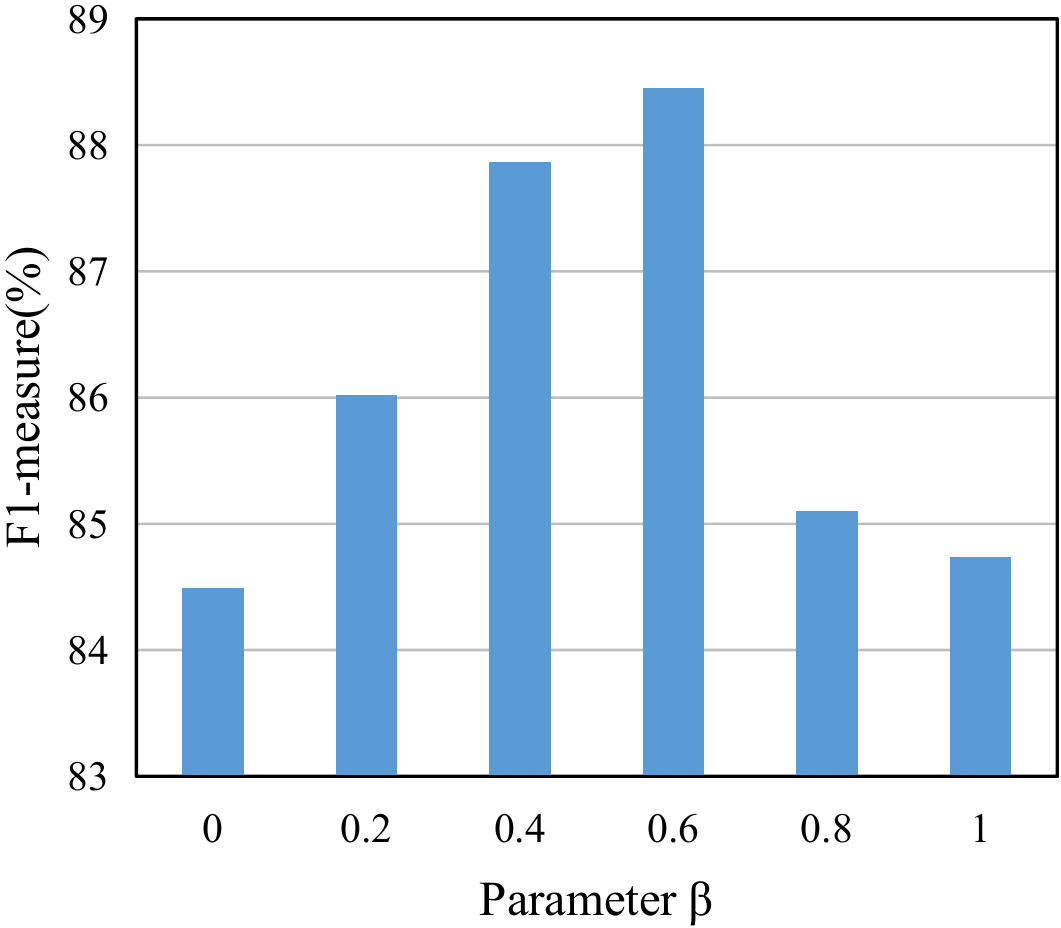}
}

\caption{The relationship of F1-measure and the parameter $\beta$ on the SZADA/1 (a) and TISZADOB/3 dataset (b).}
\label{lossfactor}
\end{figure}

\subsection{Effects of Parameter $\beta$}
The parameter $\beta$ in the loss function decides different influence of the pseudo labels generated by the traditional methods and the teacher network. To research the effects of the parameter $\beta$, we change the value of $\beta$ and obtain the corresponding change detection results. We set $\beta$ to be 0, 0.2, 0.4, 0.6, 0.8 and 1. In particular, when $\beta$ is set to be 0, only the the pseudo labels generated by the teacher network participate in the network training and when $\beta$ is set to be 1, only the the pseudo labels generated by the traditional methods participate in the network training. We show the relationship of F1-measure and the parameter $\beta$ on two datasets in Fig. \ref{lossfactor}. It can be seen that when the parameter $\beta$ is set to be 0.6, the proposed method has the best performance on the two dataset. Furthermore, if only use the pseudo labels generated by the traditional methods or the teacher network to train the student network, lower F1-measure value is obtained. This prove the usage of the jointly pseudo labels can reduces the impact of false information in the single set of pseudo labels and improve the effect of change detection.

\subsection{Effects of the Image Filter}
We design an image filter to restrain the adverse impact of the wrong information in the pseudo labels. Table \ref{filter teacher} presents the performance of the teacher network with and without the image filtering operation. On the SZADA/1 dataset, if the filter is not used, although Recall is higher, Precision and F1-measure is very low. This means a number of pixels are classified into changed by mistake. When the filter is used, the the results obtained by the proposed method can acquire a balance between Precision and Recall and obtain a higher F1-measure on the SZADA/1 dataset. On the TISZADOB/3 dataset, the three measures are all higher when the filter is used. Table \ref{filter student} presents the performance of the student network with and without the image filter operation. On the SZADA/1 dataset, Precision and F1-measure are much higher, and Recall is only a little lower when the filter is used. On the TISZADOB/3 dataset, Recall and F1-measure are higher, and Precision is a little lower when the filter is used. In general, the usage of the image filter can improve the effect of change detection effectively.

\begin{table}[]
\centering
\renewcommand{\arraystretch}{1.1}
\caption{Change Detection Result of Whether Image Filtering is Used in Teacher Network Training}
\label{filter teacher}
\begin{tabular}{ccp{1.5cm}<{\centering}p{1.5cm}<{\centering}}
\hline
                             &                           & \multicolumn{2}{c}{Using filter or not}              \\
\multirow{-2}{*}{Dataset}    & \multirow{-2}{*}{Metrics} & No                                  & Yes            \\ \hline
                             & Pr(\%)             & 19.02          & \textbf{47.10} \\
                             & Rc(\%)                & \textbf{47.04} & 31.93          \\
\multirow{-3}{*}{S1}    & F1(\%)                & 27.09          & \textbf{38.06} \\ \hline
                             & Pr(\%)             & 86.24          & \textbf{89.60} \\
                             & Rc(\%)                & 61.25          & \textbf{80.38} \\
\multirow{-3}{*}{T3} & F1(\%)                & 71.62          & \textbf{84.74} \\ \hline
\end{tabular}
\end{table}

\begin{table}[]
\centering
\renewcommand{\arraystretch}{1.1}
\caption{Change Detection Result of Whether Image Filtering is Used in Student Network Training}
\label{filter student}
\begin{tabular}{ccp{1.5cm}<{\centering}p{1.5cm}<{\centering}}
\hline
                             &                           & \multicolumn{2}{c}{Using filter or not}              \\
\multirow{-2}{*}{Dataset}    & \multirow{-2}{*}{Metrics} & No                                  & Yes            \\ \hline
                             & Pr(\%)             & 35.04          & \textbf{47.57} \\
                             & Rc(\%)                & \textbf{45.45} & 44.07          \\
\multirow{-3}{*}{S1}    & F1(\%)                & 39.57          & \textbf{45.75} \\ \hline
                             & Pr(\%)             & \textbf{82.19} & 81.77 \\
                             & Rc(\%)                & 94.30          & \textbf{96.31} \\
\multirow{-3}{*}{T3} & F1(\%)                & 87.83          & \textbf{88.45} \\ \hline
\end{tabular}
\end{table}

\subsection{Effects of the Network Structure}
In the feature extraction module of the proposed CNN network, the low-level features of the two images are extracted by the same branch, then the high-level features of the two images are extracted by two branches with the same structure but different parameters. To prove the advantages of this composite branch, we test the performance of three CNN models with different feature extraction modules, the composite branch module, the single branch module and the double branch module. Specific details of the composite branch module have been described in Table \ref{Architecture of the Change Detection Model}. The single branch module use a single branch to extract low and high level features of the two images. The double branch module use two separate branches with the same structure but different parameters to extract low and high level features of the two images. Each branch in the single and double branch module has the same structure with DeConv1-DConv5” in Table \ref{Architecture of the Change Detection Model}. The performance of each network is shown in Table \ref{Architectures of Feature Extraction Module}. The composite branch has the best performance. When using the single branch module, the different high-level semantic features of the two images are extracted in the same way, which had a bad effect on change detection. When using the double branches module, although the complexity of the network has increased, the performance of the network is not good. The more complex model pose greater challenges to the number of training data.
\begin{table}[]
\centering
\renewcommand{\arraystretch}{1.1}
\caption{Change Detection Comparison With Different Architectures of Feature Extraction Module}
\label{Architectures of Feature Extraction Module}
\begin{tabular}{ccp{1.5cm}<{\centering}p{1.5cm}<{\centering}p{1.5cm}<{\centering}}
\hline
Dataset                     & Metrics       & Single         & Double & Composite      \\ \hline
\multirow{3}{*}{S1}         & Pr(\%)    & 41.00          & 43.49  & \textbf{47.10} \\
                            & Rc(\%)    & \textbf{33.05} & 27.46  & 31.93          \\
                            & F1(\%)    & 36.60          & 33.67  & \textbf{38.06} \\ \hline
\multirow{3}{*}{T3}         & Pr(\%)    & \textbf{93.04} & 89.80  & 89.60          \\
                            & Rc(\%)    & 72.60          & 76.69  & \textbf{80.38} \\
                            & F1(\%)    & 81.56          & 82.73  & \textbf{84.74} \\ \hline
\end{tabular}
\end{table}

\section{Conclusion}
In this paper, we have proposed a novel change detection method named unsupervised self-training algorithm (USTA) for optical aerial images. The method is completely unsupervised since the whole process is based on the pseudo labels generated by the traditional methods. In our method, a designed CNN is used as the subject of training. Image filtering is used to weaken the influence of incorrect information in the pseudo labels. Moreover, the unsupervised self-training strategy is proposed. It is verified that the usage of the jointly pseudo labels can generate a more accurate change map. From the results of the experiments on the real datasets, we confirm that the proposed method have a state-of-the-art performance on the different measures compared to other related change detection methods.

\bibliographystyle{IEEEtran}
\bibliography{IEEEabrv,bare_jrnl}

\end{document}